\documentclass[journal,twoside]{IEEEtran}
\usepackage{amsfonts,bm,booktabs,cite,graphicx,hyperref,multirow,textcomp,url}
\usepackage{algorithm}
\usepackage{subfigure}
\usepackage{algpseudocode}
\usepackage{amsmath}
\usepackage{mathrsfs}
\usepackage{float}
\usepackage{color}
\usepackage{booktabs}
\usepackage{bbding}
\usepackage[flushleft]{threeparttable}

\makeatletter 
\renewcommand{\@thesubfigure}{\hskip\subfiglabelskip}
 \makeatother
\begin{document}
\title{Temporal Context Mining for Learned Video Compression}
\author{
	Xihua Sheng, 
	Jiahao Li,
	Bin Li,
	Li Li, \IEEEmembership{Member, IEEE},
	Dong Liu, \IEEEmembership{Senior Member, IEEE},
	Yan Lu\\
\thanks{This paper was received on July 11, 2022; revised on September 23, 2022; accepted on October 29, 2022; date of current version \today.\par 
X. Sheng, L. Li, and D. Liu are with the CAS Key Laboratory of Technology in Geo-Spatial Information Processing and Application System, University of Science and Technology of China, Hefei 230027, China (e-mail: xhsheng@mail.ustc.edu.cn, lil1@ustc.edu.cn, dongeliu@ustc.edu.cn).\par
J. Li, B. Li, and Y. Lu are with Microsoft Research Asia, Beijing 100080, China. (e-mail:li.jiahao@microsoft.com, libin@microsoft.com, yanlu@microsoft.com).\par
This paper was recommended by Associate Editor Singh Amit. (Corresponding author: Bin Li.)\par
The work was done when Xihua Sheng was a full-time intern with Microsoft Research Asia.

}
}

\markboth{IEEE TRANSACTIONS ON MultiMedia}{Sheng \MakeLowercase{\textit{et al.}}: Temporal Context Mining for Learned Video Compression}

\maketitle
\begin{abstract}
Applying deep learning to video compression has attracted increasing
attention in recent few years. In this work, we address end-to-end learned video compression with a special focus on better learning and utilizing temporal contexts. We propose to propagate not only the last reconstructed frame but also the feature before obtaining the reconstructed frame for temporal context mining. From the propagated feature, we learn multi-scale temporal contexts and re-fill the learned temporal contexts into the modules of our compression scheme, including the contextual encoder-decoder, the frame generator, and the temporal context encoder. We discard the parallelization-unfriendly auto-regressive entropy model to pursue a more practical encoding and decoding time. Experimental results show that our proposed scheme achieves a higher compression ratio than the existing learned video codecs. Our scheme also outperforms x264 and x265 (representing industrial software for H.264 and H.265, respectively) as well as the official reference software for H.264, H.265, and H.266 (JM, HM, and VTM, respectively). Specifically, when intra period is 32 and oriented to PSNR, our scheme outperforms H.265--HM by 14.4\% bit rate saving; when oriented to MS-SSIM, our scheme outperforms H.266--VTM by 21.1\% bit rate saving.
\end{abstract}
\IEEEpeerreviewmaketitle

\section{Introduction}
\label{sec:intro}
Video data contributes to most of the internet traffic nowadays. Therefore, efficient video compression is always a high demand to reduce the transmission and storage cost. In the past two decades, several video coding standards have been developed, including H.264/AVC~\cite{wiegand2003overview},  H.265/HEVC~\cite{sullivan2012overview}, and H.266/VVC~\cite{bross2021overview}.\par
\begin{figure}[htb]
  \centering
   \includegraphics[height=4.5cm]{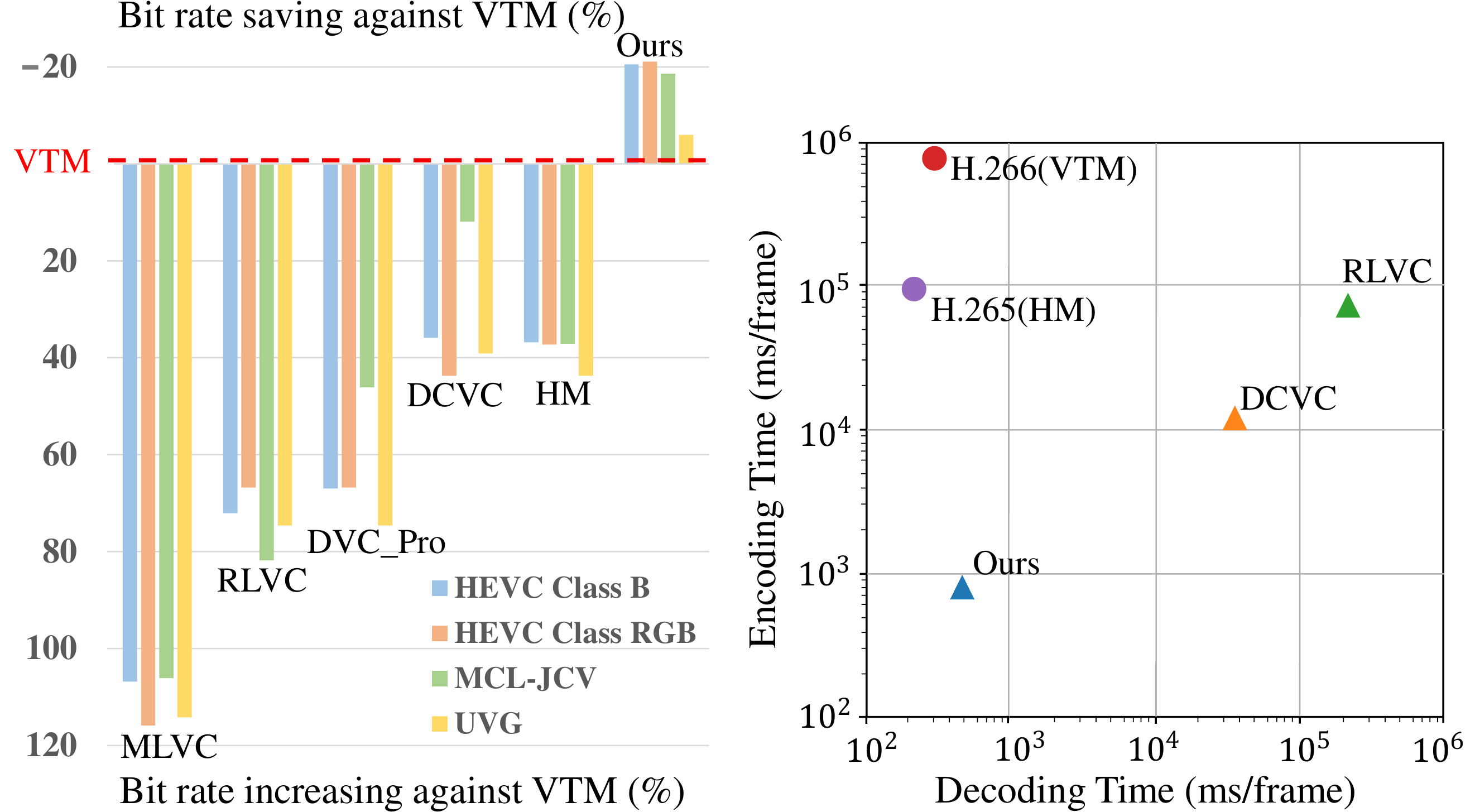}
   \caption{Compression ratio in terms of MS-SSIM when intra period is 32 and running
time on 1080p videos. As traditional video codecs are optimized for CPU, we report their run time on CPU, which is the same as previous schemes\cite{lu2020end,Rippel_2021_ICCV}. The learned video codecs are run on GPU.}
   \label{fig:decoding_time}
\end{figure}
Recently, learned video compression has explored a new direction. 
Existing learned video compression can be roughly categorized into four classes.
Residual coding-based schemes~\cite{liu2020learned,Rippel_2021_ICCV,hu2020improving,lu2020content,lu2020end,lin2020m,hu2021fvc,yang2020learning,agustsson2020scale,cheng2019learning,rippel2019learned,djelouah2019neural,yang2021learning,wu2018video,liu2021deep,liu2020neural,yilmaz2021end,chen2019learning,ma2019image}: prediction is first applied in the pixel domain or feature domain. Then the residue from the prediction is compressed.
Volume coding-based schemes~\cite{Habibian_2019_ICCV,sun2020high,pessoa2020end}: videos are regarded as volumes containing multiple frames. 3D convolutions are applied to get the latent representations. Entropy coding-based schemes~\cite{liu2020conditional}: each frame is encoded with independent image codec. The correlations between the latent representations in different frames are explored to guide the entropy modeling.
Conditional coding-based schemes~\cite{li2021deep}: temporal contexts are served as conditions and the temporal correlation is explored by the encoder automatically. There are also some other schemes focusing on the perceptual quality optimization for
learned codecs~\cite{wang2020multi,ijcai2022p214}. We agree that perceptual quality is more important and learned codecs have benefits on optimizing for perceptual quality over traditional codecs due to the end-to-end training and back-propagation. However, this paper still focuses on objective distortion metrics to make the comparison with previous schemes easier. \par
Among the existing learned video codecs, Li~\emph{et al.} proposed a deep contextual video compression (DCVC) framework and achieved the state-of-the-art compression ratio~\cite{li2021deep}. DCVC is an open framework based on the aforementioned conditional coding structure. DCVC generates a single-scale temporal context from the previously decoded frame with motion compensation and several convolutional layers. It is a simple yet useful method. However, the previously decoded frame loses much texture and motion information as it only contains three channels. Moreover, a single-scale context may not contain sufficient temporal information. Therefore, in this paper, we try to find a better way to learn and utilize temporal contexts.\par
Following the conditional coding structure, we propose a temporal context mining (TCM) module to learn richer and more accurate temporal contexts. We propagate the feature before obtaining the reconstructed frame and learn temporal contexts from the propagated features for the current frame using the TCM module. Considering that learning a single-scale context may not describe the spatio-temporal non-uniformity well~\cite{ding2007adaptive,Wu2001AFF,ma2020end}, the TCM module adopts a hierarchical structure to generate multi-scale temporal contexts. Finally, we re-fill the learned temporal contexts into the modules of our compression scheme, including the contextual encoder-decoder, the frame generator, and the temporal context encoder. We refer to the procedure as temporal context re-filling (TCR).\par

Following most previous schemes, this paper focuses on the scenarios with delay constraint or called low-delay coding although random-access is the most efficient coding configuration. Some schemes reported higher compression ratios than H.265/HEVC, but most of them just used x265 \cite{x265}, which is optimized for fast encoding time and cannot represent the best compression ratio of H.265/HEVC. In this paper, we try our best to compare with the best traditional video codecs. We use the official reference software JM~\cite{JM}, HM~\cite{HM}, and VTM~\cite{VTM} for H.264/AVC, H.265/HEVC, and H.266/VVC to represent the highest potential compression ratio, without playing any tricks to harm it. More details could be found in Sec.~\ref{comparison_with_traditional_video}. In addition, many schemes~\cite{wu2018video,lu2020end,Habibian_2019_ICCV,li2021deep,DBLP:conf/iclr/YangYMM21} used the auto-regressive entropy model~\cite{DBLP:conf/nips/MinnenBT18} for a higher compression ratio, which poses significant challenges for parallel implementation and meanwhile greatly increases the decoding time. We do not apply the auto-regressive entropy model in any part of our scheme to build a parallelization-friendly decoder, although we agree that the auto-regressive entropy model could further boost the compression ratio by a large margin. Fig.~\ref{fig:decoding_time} compares the compression ratio and running time on 1080p videos. Our scheme achieves much faster encoding and decoding speed than the other learned video codecs. To the best of our knowledge, this is the first end-to-end learned video compression scheme achieving such a milestone result that outperforms HM in terms of peak signal-to-noise ratio (PSNR) and VTM in terms of multi-scale structural similarity index measure (MS-SSIM)~\cite{wang2003multiscale}.\par

Our contributions are summarized as follows:
\begin{itemize}
    \item We propose a temporal context mining module to learn multi-scale temporal contexts from the propagated features rather than the previously reconstructed frames.
    \item We propose to re-fill the learned multi-scale temporal contexts into the contextual encoder-decoder, the frame generator, and the temporal context encoder to help compress and reconstruct the current frame.
    \item Without the auto-regressive entropy model, our proposed scheme achieves higher compression ratio than the existing learned video codecs. Our scheme also outperforms the reference software of H.265/HEVC---HM by 14.4\% in terms of PSNR and outperforms the reference software of H.266/VVC---VTM by 21.1\% in terms of MS-SSIM. We will release code to facilitate the future investigation.
\end{itemize}
The remainder of this paper is organized as follows. Section~\ref{sec:related_work} gives a brief review of related work. Section~\ref{sec:methodology} describes the methodology of our proposed scheme. Section~\ref{sec:experiments} presents the experimental results and analysis of the proposed scheme.  Section~\ref{sec:conclusion} concludes this paper and discuss the future investigate. 
\section{Related Work}
\label{sec:related_work}
\subsection{Traditional Video Compression}
Traditional video compression has been developed for several decades, and several video coding standards have been proposed. H.264/AVC~\cite{sullivan2012overview} was initially developed in the period between 1999 and 2003 by the well-known ITU-T and ISO/IEC standards. It achieves great success and is widely used for many applications, such as broadcast of high-definition TV signals and internet and mobile network video. Along with the increase of video resolution and use of parallel processing architectures, H.265/HEVC~\cite{wiegand2003overview} was finalized in 2013 and offered about 50\% bit rate saving over H.264/AVC. The superior compression efficiency of H.265/HEVC enables the popularity of 4K video with increased fidelity. H.266/VVC~\cite{bross2021overview} is the new generation of international video coding standard. It is designed not only to reduce a substantial bit rate compared to H.265/HEVC, but also to cover all current and emerging media needs. These video coding standards follow a similar hybrid video coding framework, including prediction, transform, quantization, entropy coding, and loop filtering. Although the learned image codecs~\cite{guo2021causal,cheng2020image} have caught up with or even surpassed the traditional image coding schemes, the traditional video compression scheme is still state-of-the-art in terms of compression ratio.
\subsection{Learned Video Compression}
Learned video compression has explored a new direction in recent years.
Lu~\emph{et al.} replaced each part in the traditional motion-compensated prediction and residual coding framework with convolutional neural networks~\cite{lu2020end}. They jointly optimized the whole network with the rate-distortion cost. Lin~\emph{et al.} explored multiple reference frames and associated multiple motion vectors to generate a more accurate prediction of the current frame and reduce the coding cost of motion vectors \cite{lin2020m}. Yang~\emph{et al.} designed a recurrent learned video compression scheme \cite{yang2021learning}. The proposed recurrent auto-encoder and recurrent probability model use the temporal information in a large range of frames to generate latent representations and reconstruct compressed output. Hu~\emph{et al.} extracted the features of the input frames and applied deformable convolution to perform motion prediction~\cite{hu2021fvc}.  The learned offset and residue are compressed in the feature domain. Then multiple reference features stored in the decoded buffer are fused by a non-local attention module to obtain the reconstructed frame. Except for the motion-compensated prediction and residual coding framework, Habibian~\emph{et al.} regarded the video as a volume of multiple frames and proposed a 3D auto-encoder to directly compress multiple frames~\cite{Habibian_2019_ICCV}. Liu~\emph{et al.} used the image codec to compress each frame and proposed an entropy model to explore the temporal correlation of the latent representations~\cite{liu2020conditional}. Li~\emph{et al.} shifted the paradigm from residual coding to conditional coding~\cite{li2021deep}. They learned temporal contexts from the previously decoded frame and let the encoder explore the temporal correlation to remove the redundancy automatically.
\begin{figure*}[htb]
  \centering
   \includegraphics[width=0.9\linewidth]{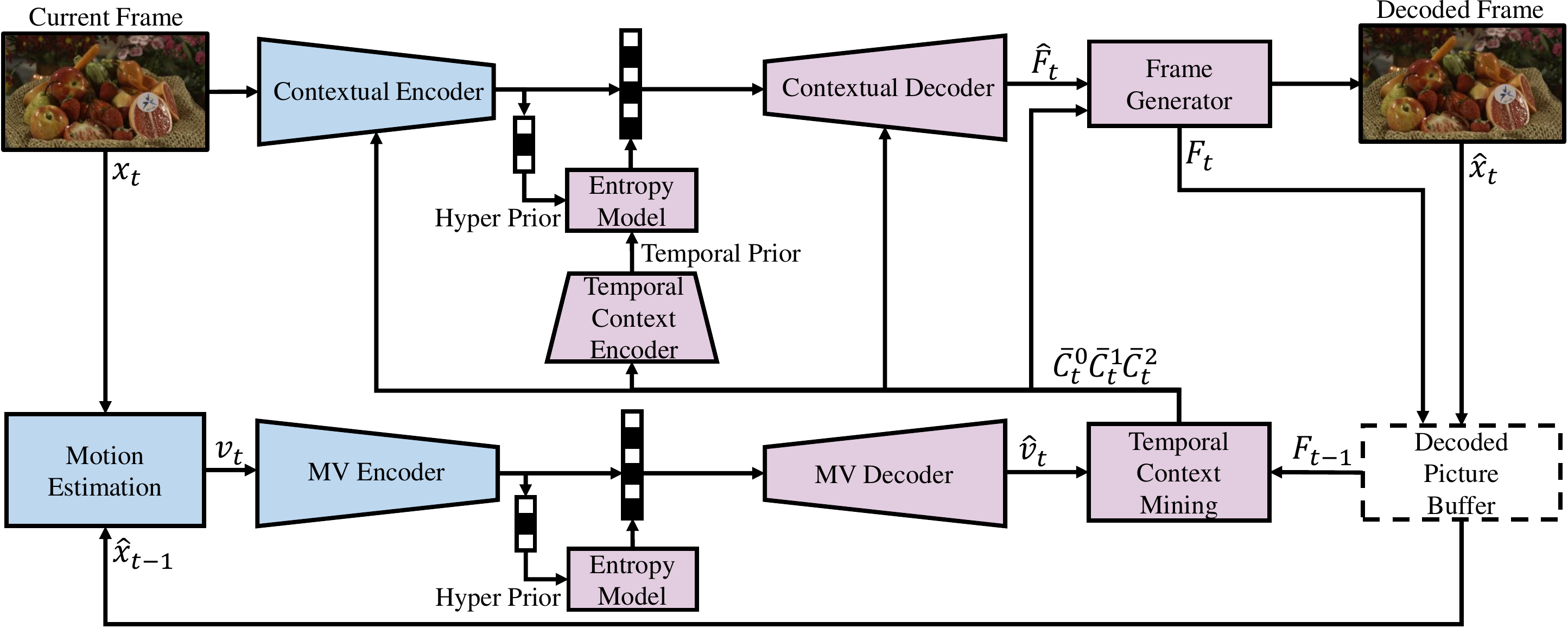}
   \caption{Overview of our proposed video compression scheme. 
   Given an input frame $x_t$, the motion vector (MV) between $x_t$ and previously decoded frame $\hat{x}_{t-1}$ is estimated, compressed, and then reconstructed. A temporal context mining (TCM) module is proposed to learn multi-scale temporal contexts from the propagated feature $F_{t-1}$ instead of the previously decoded frame $\hat{x}_{t-1}$. The temporal contexts ($\bar{C}_t^{0},\bar{C}_t^{1},\bar{C}_t^{2}$) are re-filled into the contextual encoder-decoder, the frame generator, and the temporal context encoder to compress and reconstruct the current frame $x_t$. The decoded frame $\hat{x}_t$ and the feature $F_t$ before obtaining $\hat{x}_t$ are propagated to help compress the next frame $x_{t+1}$.
   The blue modules are only used at the encoder side. The lavender modules are used at both the encoder and the decoder sides.}
   \label{fig:framework}
\end{figure*}
\subsection{Traditional Video Codecs versus Learned Video Codecs}\label{comparison_with_traditional_video}
Although many existing learned video codecs reported a better compression ratio than H.265/HEVC, they may not use the best encoder with the following major issues. Firstly, x264~\cite{x264} and x265~\cite{x265} are used to represent the compression ratio of H.264/AVC and H.265/HEVC, respectively. However, they are optimized for faster encoding instead of a higher compression ratio. The official reference software JM~\cite{JM}, HM~\cite{HM}, and VTM~\cite{VTM} should be used to represent the higher compression ratio. For example, the gap between x265 and HM is about 80\%, as shown in Table~\ref{table:ip32_psnr}.
Secondly, an unrealistic intra period of 10 or 12 is used to limit temporal error propagation~\cite{mentzer2022neural}. However, in real applications, the bits of I frames account for a substantial part of the total number of bits when using such a small intra period. Taking the VTM as an example, the bits of I frame account for 51\% for intra period 12 and account for 29\% for intra period 32. Intra period 12 is harmful to the compression ratio and is seldom used in the real world. Therefore, this paper proposes to use intra period 32 (roughly 1 second for 30fps), which is more reasonable to measure the performance of inter coding for video codecs. Thirdly, the internal color space of YUV420 is set for traditional codecs in existing schemes. However, after comparing different color spaces for traditional codecs, we conclude that YUV444 is better than YUV420 and RGB, even though the final distortion is measured in RGB. The performance gap is more than 30\%. Thus, we propose to use YUV444 as the internal color space for JM, HM, and VTM in favor of seeking the higher potential compression ratio of the traditional codecs. A similar conclusion has been made in learned image compression~\cite{guo2021causal}. 
Fourthly, we use the best suitable configuration (e.g., using range extension profile to encode YUV444 content) for traditional codecs as the performance gap of different configurations can be over 40\%.\par

\section{Methodology}
\label{sec:methodology}
\subsection{Overview}\label{baseline}
Aiming to find a better way to learn and utilize temporal contexts, we propose a new learned video compression scheme based on temporal context mining and re-filling. An overview of our scheme is depicted in Fig.~\ref{fig:framework}.\par
\subsubsection{Motion Estimation} We feed the current frame $x_t$ and the previously decoded frame $\hat{x}_{t-1}$ into a neural network-based motion estimation module to estimate the optical flow. The optical flow is considered as the estimated motion vector (MV) $v_t$ for each pixel. In this paper, the motion estimation module is based on the pre-trained Spynet~\cite{ranjan2017optical}.

\subsubsection{MV Encoder-Decoder} After obtaining the motion vectors $v_t$, we use the MV 
encoder and decoder to compress and reconstruct the input MV $v_t$ in a lossy way. Specifically, $v_t$ is compressed by an auto-encoder with the hyper prior structure~\cite{DBLP:conf/iclr/BalleMSHJ18}. The reconstructed MV is denoted as $\hat{v}_t$.

\subsubsection{Learned Temporal Contexts} We propose a TCM module to learn richer
and more accurate temporal contexts from the propagated feature $F_{t-1}$ instead of the previously decoded frame $\hat{x}_{t-1}$. Instead of producing only a single-scale temporal context, the TCM module generates multi-scale temporal contexts $\bar{C_t}^{l}$ to capture spatial-temporal non-uniform motion and texture, where $l$ is the index of different scales. The learned temporal contexts are re-filled in the contextual
encoder-decoder, the frame generator, and the temporal context
encoder to help improve the compression ratio. This procedure is referred to as TCR. We will introduce them in detail in Section~\ref{TCM} and~\ref{TCR}. 

\subsubsection{Contextual Encoder-Decoder and Frame Generator} With the assistance of the re-filled multi-scale temporal contexts $\bar{C_t}^{l}$, the contextual encoder-decoder and the frame generator are used to compress and reconstruct the current frame $x_t$. We denote the decoded frame as $\hat{x}_t$ and the feature before obtaining $\hat{x}_t$ as $F_t$.  $\hat{x}_t$ and $F_t$ are propagated to help compress the next frame $x_{t+1}$. The details are presented in Section~\ref{TCR}.

\subsubsection{Temporal Context Encoder}
To utilize the temporal correlation of the latent representations of different frames produced by the contextual encoder, we use a temporal context encoder to generate the temporal prior by taking advantage of the multi-scale temporal contexts $\bar{C_t}^{l}$. More information is provided in Section~\ref{TCR}.

\subsubsection{Entropy Model}
We use the factorized entropy model for hyper prior and Laplace distribution to model the latent representations as \cite{li2021deep}. We do not apply the auto-regressive entropy model to make the decoding processes parallelization-friendly, even though it is helpful to improve the compression ratio. For the contextual encoder-decoder, we fuse the hyper and temporal priors~\cite{li2021deep} to estimate more accurate parameters of Laplace distribution. When writing bitstream, we apply a similar implementation as in \cite{begaint2020compressai}.
\begin{figure}
  \centering
   \includegraphics[width=\linewidth]{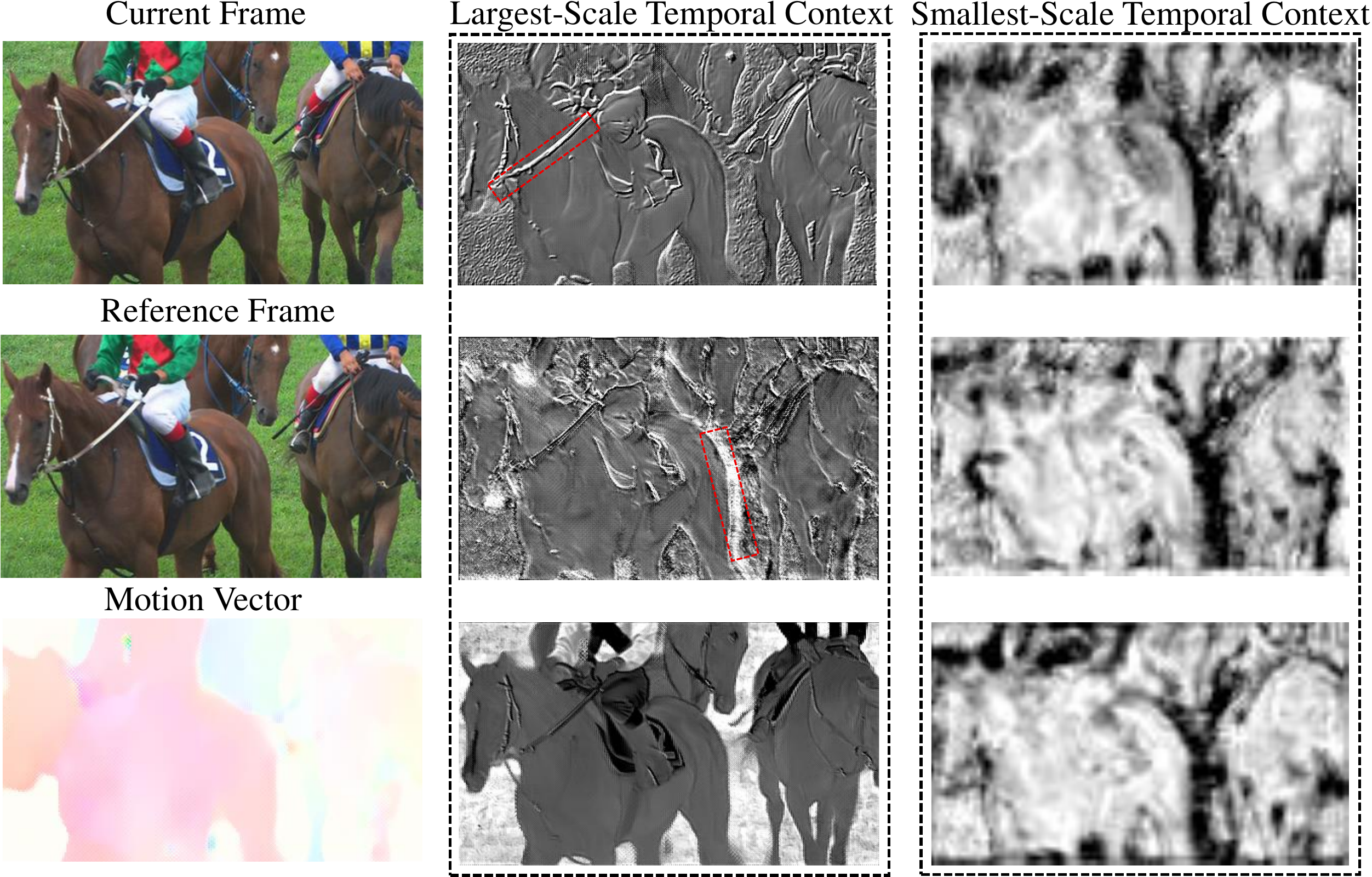}
   \caption{Visualization of the multi-scale temporal contexts. In the largest-scale context, some channels focus on the texture information and some focus on the color information. In the smallest-scale context, channels mainly focus on the regions with large motion.}
   \label{fig:visualize_context}
\end{figure}

\subsection{Temporal Context Mining}\label{TCM}
Considering that the previously decoded frame $\hat{x}_{t-1}$ loses much information as it only contains three channels, it is not optimal to learn temporal contexts from $\hat{x}_{t-1}$.  In our paper, we propose a TCM module to learn temporal contexts from the propagated feature $F_{t-1}$. Different from the existing video compression schemes in feature domain~\cite{hu2021fvc,feng2020learned, yang2020learning}, which extract features from the previously decoded frame $\hat{x}_{t-1}$, we propagate the feature $F_{t-1}$ before obtaining $\hat{x}_{t-1}$. Specifically, in the reconstruction procedure of $\hat{x}_{t-1}$, we store the feature $F_{t-1}$ before the last convolutional layer of the frame generator in the generalized decoded picture buffer (DPB). To reduce the computational complexity, instead of storing multiple features of previously decoded frames~\cite{hu2021fvc, yang2020learning}, we only store a single feature. Then $F_{t-1}$ is propagated to learn temporal contexts for compressing the current frame $\hat{x}_{t}$. For the first P frame, we still extract the feature from the reconstructed I frame, to make the model adapt to different image codecs. \par

Besides, learning a single scale context may not describe the spatio-temporal non-uniform motion and texture well~\cite{ding2007adaptive,Wu2001AFF,ma2020end}. As shown in Fig.~\ref{fig:visualize_context}, in the largest-scale context, some channels focus on the texture information and some focus on the color information. In the smallest-scale context, channels mainly focus on the regions with large motion. Therefore, a hierarchical approach is performed to learn multi-scale temporal contexts. \par
\begin{figure}[t]
  \centering
   \includegraphics[width=\linewidth]{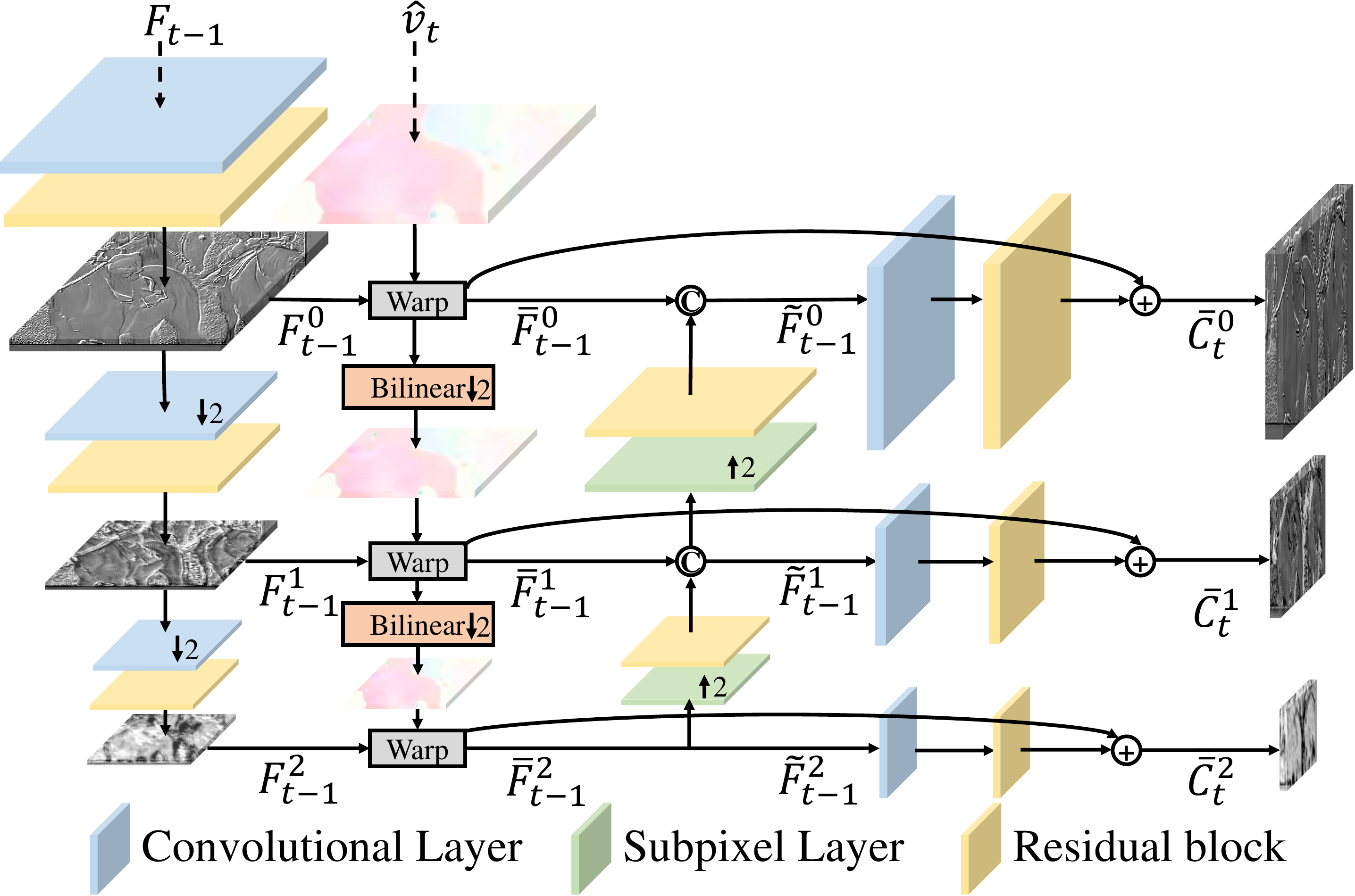}
   \caption{Architecture of the temporal context mining (TCM) module. The propagated feature $F_{t-1}$ is fed into the TCM module to generate multi-scale temporal contexts $\bar{C_t}^{l}$.}
   \label{fig:temporal_context_mining}
\end{figure}

As shown in Fig.~\ref{fig:temporal_context_mining}, we first generate multi-scale features $F_{t-1}^{l}$ from the propagated feature $F_{t-1}$ using a feature extraction module ($extract$) with $L$ levels which consists of convolutional layers and residual blocks~\cite{he2016deep} (three levels are used in our paper).
\begin{equation}
F_{t-1}^{l}=extract\left(F_{t-1}\right), l=0,1,2
\label{equ1}
\end{equation}
Meanwhile, the decoded MV $\hat{v}_t$ are downsampled using the bilinear filter to generate multi-scale MVs $\hat{v}_{t}^{l}$, where $\hat{v}_{t}^{0}$ is set to $\hat{v}_t$. Note that each downsampled MV is divided by 2. Then, we warp ($warp$) the multi-scale features ${F}_{t-1}^{l}$ using the associated MV $\hat{v}_{t}^{l}$ at the same scale.
\begin{equation}
\bar{F}_{t-1}^{l}=warp\left(F_{t-1}^{l}, \hat{v}_{t}^{l}\right), l=0,1,2
\label{equ3}
\end{equation}
After that, we use an upsample ($upsample$) module, consisting of one subpixel layer~\cite{shi2016real} and one residual block, to upsample $\bar{F}_{t-1}^{l+1}$. The upsampled feature is then concatenated ($concat$) with $\bar{F}_{t-1}^{l}$ at the same scale.
\begin{equation}
\tilde{F}_{t-1}^{l} = concat\left(\bar{F}_{t-1}^{l}, upsample\left(\bar{F}_{t-1}^{l+1}\right)\right), l=0,1
\label{equ4}
\end{equation}
At each level of the hierarchical structure, a context refinement module $(refine)$, consisting of one convolutional layer and one residual block, is used to learn the residue~\cite{zhang2017beyond}. The residue is added to $\bar{F}_{t-1}^{l}$ to generate the final temporal contexts $\bar{C}_{t}^{l}$, as illustrated in Fig~\ref{fig:temporal_context_mining}.
\begin{equation}
\bar{C}_{t}^{l} = \bar{F}_{t-1}^{l}+ refine\left(\tilde{F}_{t-1}^{l}\right), l=0,1,2
\label{equ5}
\end{equation}
\begin{figure}[t]
  \centering
    \includegraphics[height=8.0cm]{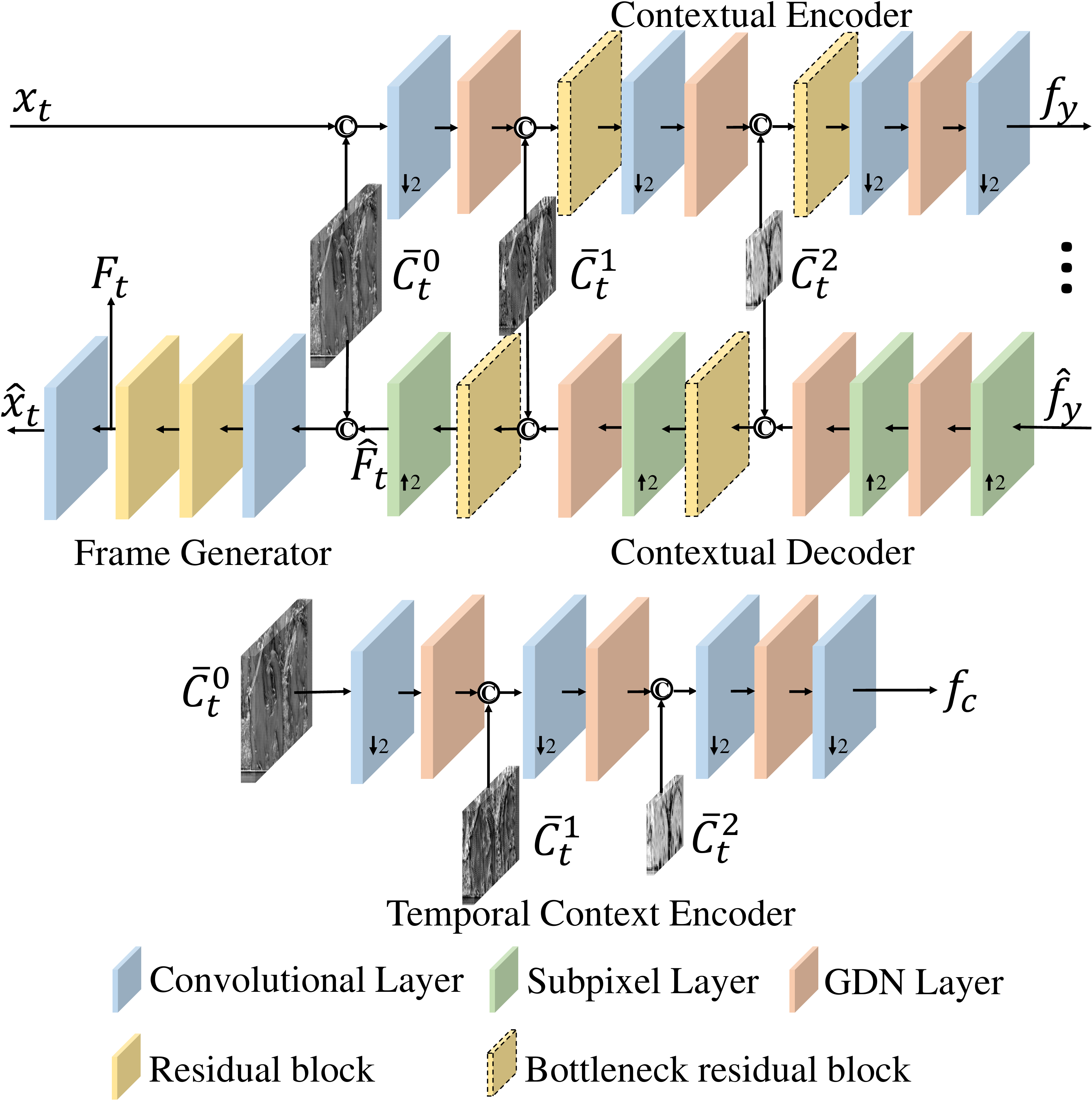}
\caption{Illustration of temporal context re-filling (TCR). The multi-scale temporal contexts are re-filled into the contextual encoder-decoder, frame generator, and temporal context encoder.}
\label{fig:temproal_context_refilling}
\end{figure}
\begin{figure}[t]
  \centering
    \includegraphics[height=6.5cm]{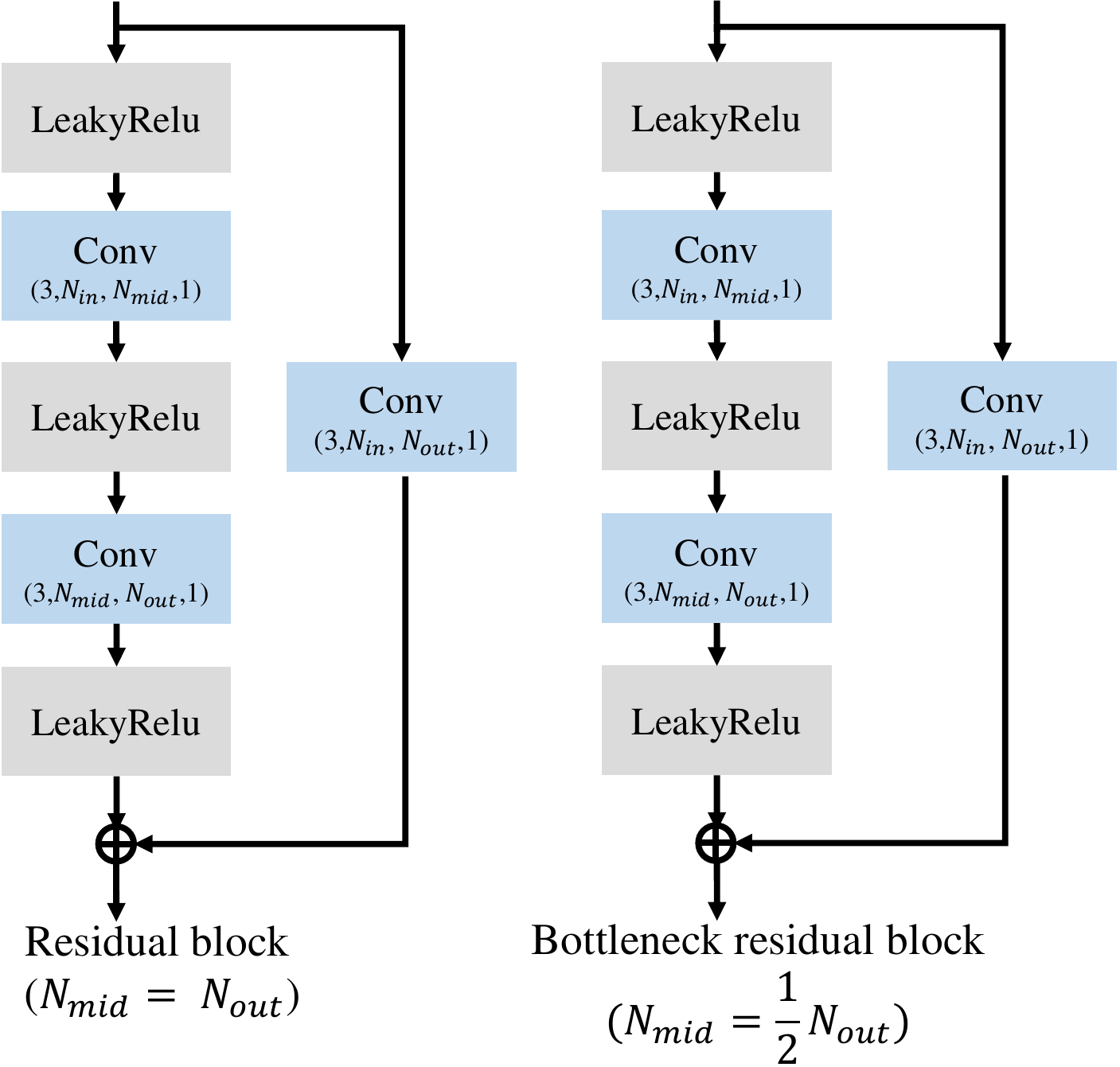}
\caption{Architectures of residual block and bottleneck residual block. The numbers in a convolutional layer like ($3, N_{in}, N_{mid}, 1$) refer to the kernel size is 3, the number of input channels is $N_{in}$, the number of output channels is $N_{mid}$, and the stride is 1.}
\label{fig:resblock}
\end{figure}
\subsection{Temporal Context Re-filling}\label{TCR}
To fully take advantage of the temporal correlation, we re-fill the learned multi-scale temporal contexts into the modules of our compression scheme, including the contextual encoder-decoder, the frame generator, and the temporal context encoder, as shown in Fig.~\ref{fig:temproal_context_refilling}. The temporal context plays an important role in temporal prediction and temporal entropy modeling. With the re-filled multi-scale temporal contexts, the compression ratio of our scheme is improved a lot.\par
\subsubsection{Contextual Encoder-Decoder and Frame Generator} We concatenate the largest-scale temporal context $\bar{C}_{t}^{0}$ with the current frame $x_t$, and then feed them into the contextual encoder. In the process of mapping from $x_t$ to the latent representation $f_y$, we also concatenate $\bar{C}_{t}^{1}$ and $\bar{C}_{t}^{2}$ with other scales into the encoder. Symmetric with the encoder, the contextual decoder maps the quantized latent representation $\hat{f}_y$ to the feature $\hat{F}_t$ with the assistance of $\bar{C}_{t}^{1}$ and $\bar{C}_{t}^{2}$. Then $\hat{F}_t$ and $\bar{C}_{t}^{0}$ are concatenated and fed into the frame generator to obtain the reconstructed frame $\hat{x}_t$. Considering that the concatenation increases the number of channels, a ``bottleneck" building residual block is used to reduce the complexity of the middle layer. The detailed architectures are illustrated in Fig.~\ref{fig:resblock}. The feature $F_t$ before the last convolutional layer of the frame generator is propagated  to help compress the next frame $x_{t+1}$.\par
\subsubsection{Temporal Context Encoder} 
To explore the temporal correlation of the latent representations of different frames, we use a temporal context encoder to obtain the lower-dimensional temporal prior $f_c$. Instead of using a single-scale temporal context as~\cite{li2021deep}, we concatenate the multi-scale temporal contexts in the process of generating temporal prior. The temporal prior is fused with the hyper prior to estimate the means and variance of Laplacian distribution for the latent representation $\hat{f}_y$. The architecture of the temporal context encoder is presented in Fig.~\ref{fig:temproal_context_refilling}.\par
\subsection{Loss Function}\label{loss}
Our scheme targets to jointly optimize the rate-distortion (R-D) cost.
\begin{equation}
L_t=\lambda D_t + R_t= \lambda d(x_t,\hat{x}_t) + R_{t}^{\hat{v}} +R_{t}^{\hat{f}}
\label{loss}
\end{equation}
$L_t$  is the loss function for the current time step $t$.
$d(x_t,\hat{x}_t)$ refers to the distortion between the input frame $x_t$ and the reconstructed frame $\hat{x}_t$, where $d(\cdot)$ denotes the mean-square-error or 1$-$MS-SSIM~\cite{wang2003multiscale}. $R_{t}^{\hat{v}}$ represents the bit rate used for encoding the quantized motion vector latent representation and the associated hyper prior. $R_{t}^{\hat{f}}$ represents the bit rate used for encoding the quantized contextual latent representation and the associated hyper prior. We train our model step-by-step to make the training more stable as previous the scheme\cite{lin2020m}. It is worth mentioning that, in the last five epochs, we use a commonly-used training strategy in recent papers~\cite{lu2020content,yang2021learning,mentzer2022neural,guo2021learning}, that trains our model on the sequential training frames to alleviate the error propagation.
\begin{equation}
L^{T}=\frac{1}{T} \sum_{t} L_{t}=\frac{1}{T} \sum_{t}\left\{\lambda d\left(x_{t}, \hat{x}_{t}\right)+R_{t}^{\hat{v}} +R_{t}^{\hat{f}}\right\}
\label{cascaded_loss}
\end{equation}
where $T$ is the time interval and set as 4 in our experiments.
\section{Experiments}
\label{sec:experiments}
\subsection{Experimental Setup}
\label{sec:experimental_setup}
\begin{figure}[t]
  \centering
   \includegraphics[width=\linewidth]{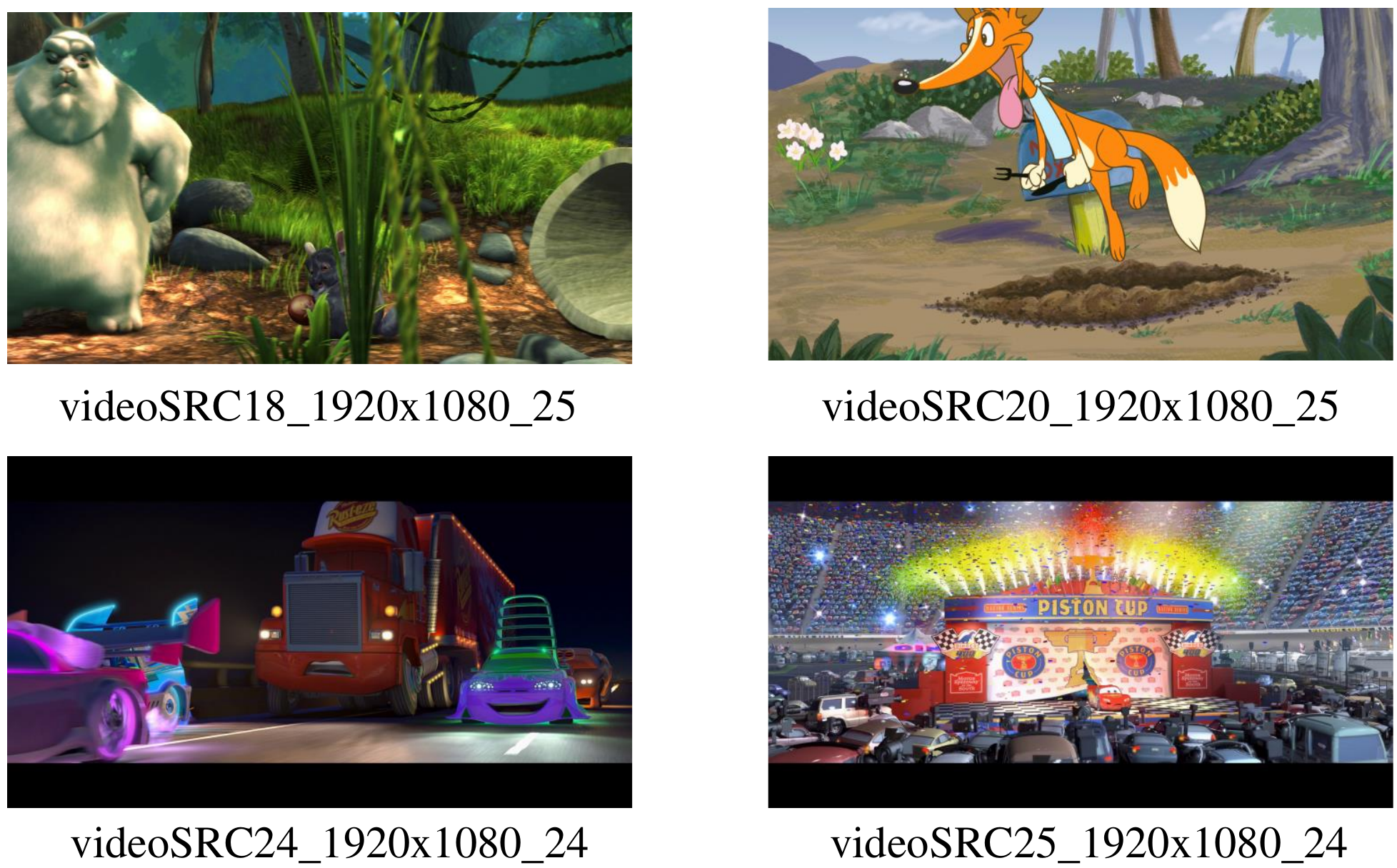}
   \caption{Visualization of the four animations sequences (18, 20, 24, 25) which are excluded by the MCL-JCV-26 dataset.}
   \label{fig:MCL-JCV-26}
\end{figure}
\begin{figure}[t]
  \centering
   \includegraphics[width=\linewidth]{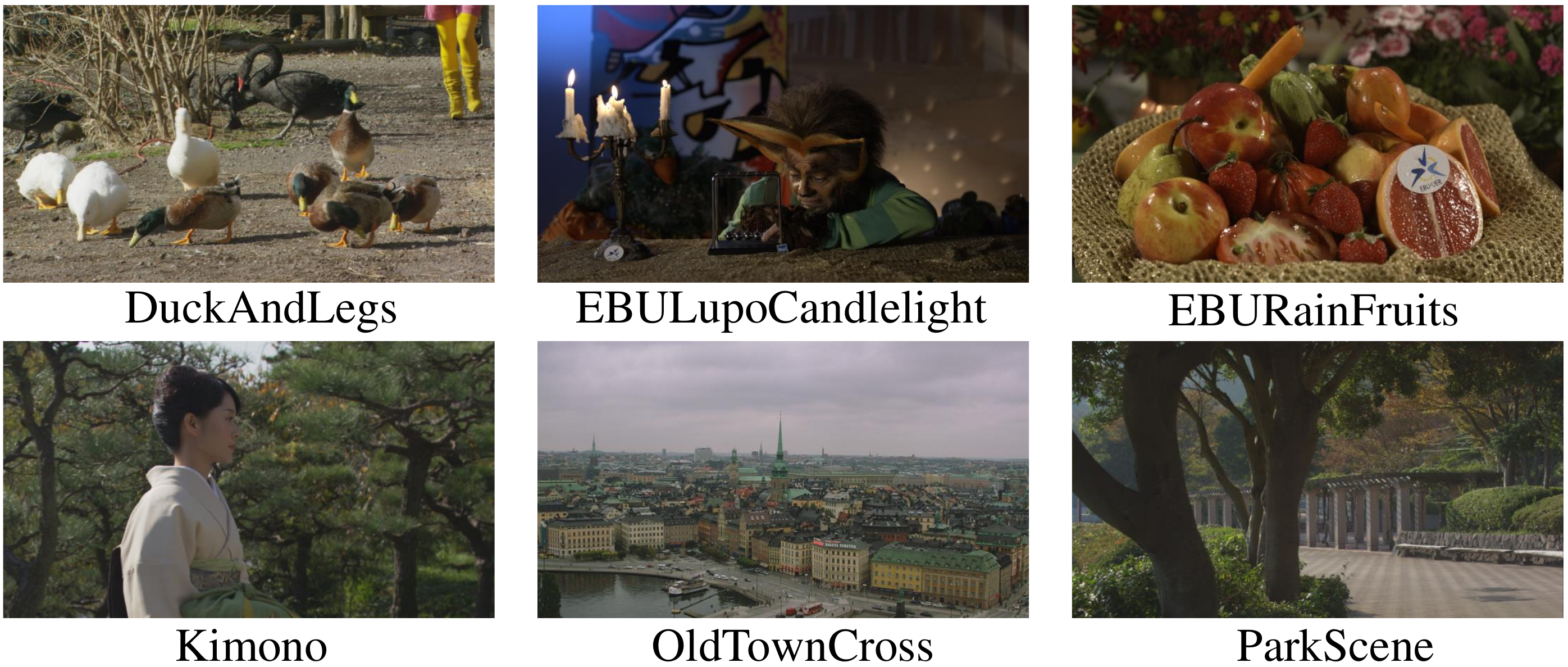}
   \caption{HEVC Class RGB consists of 6 1080p video sequences with 10 bit depth and in RGB444 format defined in the common test conditions for HEVC range extensions~\cite{flynn2015overview}, including \emph{DucksAndLegs}, \emph{EBULupoCandlelight}, \emph{DucksAndLegs}, \emph{DucksAndLegs}, \emph{Kimono}, \emph{OldTownCross}, and \emph{ParkScene}.}
   \label{fig:Class RGB}
\end{figure}
\subsubsection{Datasets} Many training datasets are produced for learned video compression~\cite{ma2021bvi,katsenou2020bvi,xue2019video}. Following most previous schemes\cite{lin2020m,lu2020end,yang2021learning,li2021deep}, in our work, we use the commonly-used Vimeo-90k~\cite{xue2019video} training split, and randomly crop the videos into $256 \times 256$ patches. To evaluate the performance and application domain of our proposed video compression scheme, we use UVG~\cite{mercat2020uvg}, MCL-JCV~\cite{wang2016mcl}, and HEVC datasets~\cite{sullivan2012overview}. These datasets contain various video contents, including slow/fast motion and poor/high quality, which are commonly used in leaned video compression.
The UVG dataset has 7 1080p sequences and the MCL-JCV dataset has 30 1080p sequences. Considering the animations sequences in MCL-JCV (18, 20, 24, 25) are quite different from the natural videos in the training dataset, as shown in Fig.~\ref{fig:MCL-JCV-26}, we follow the setting in~\cite{agustsson2020scale} and build another MCL-JCV-26 dataset which excludes the four animations sequences. The HEVC dataset contains 16 sequences including Class B, C, D, and E.  The source videos of these datasets are in YUV420 format while PSNR is calculated in RGB color space.  Therefore, we build an additional class called HEVC Class RGB, which includes add 6 1080p videos with 10 bit depth and in RGB444 format defined in the common test conditions for HEVC range extensions~\cite{flynn2015overview}: \emph{DucksAndLegs}, \emph{EBULupoCandlelight}, \emph{DucksAndLegs}, \emph{DucksAndLegs}, \emph{Kimono}, \emph{OldTownCross}, and \emph{ParkScene},  as illustrated in Fig.~\ref{fig:Class RGB}. \par
\subsubsection{Implementation Details}
We train four models with different $\lambda$ values ($\lambda=256, 512, 1024, 2048$) for multiple coding rates. The AdamW~\cite{DBLP:conf/iclr/LoshchilovH19} optimizer is used and the batch size is set to 4. When using MS-SSIM for performance evaluation, we fine-tune the model by using 1$-$MS-SSIM as the distortion loss with different $\lambda$ values ($\lambda=8, 16, 32, 64$). Our model is implemented with PyTorch and trained on 2 NVIDIA V100 GPUs. It takes 2.5 days to train our model.\par
\subsubsection{Evaluation Metrics}
We use bpp (bit per pixel) to measure the bits cost for one pixel in each frame. We use PSNR and MS-SSIM to evaluate the distortion between the decoded frame and the original frame.
\begin{figure*}[t]
  \centering
  \begin{minipage}[c]{0.29\linewidth}
  \centering
  \includegraphics[width=\linewidth]{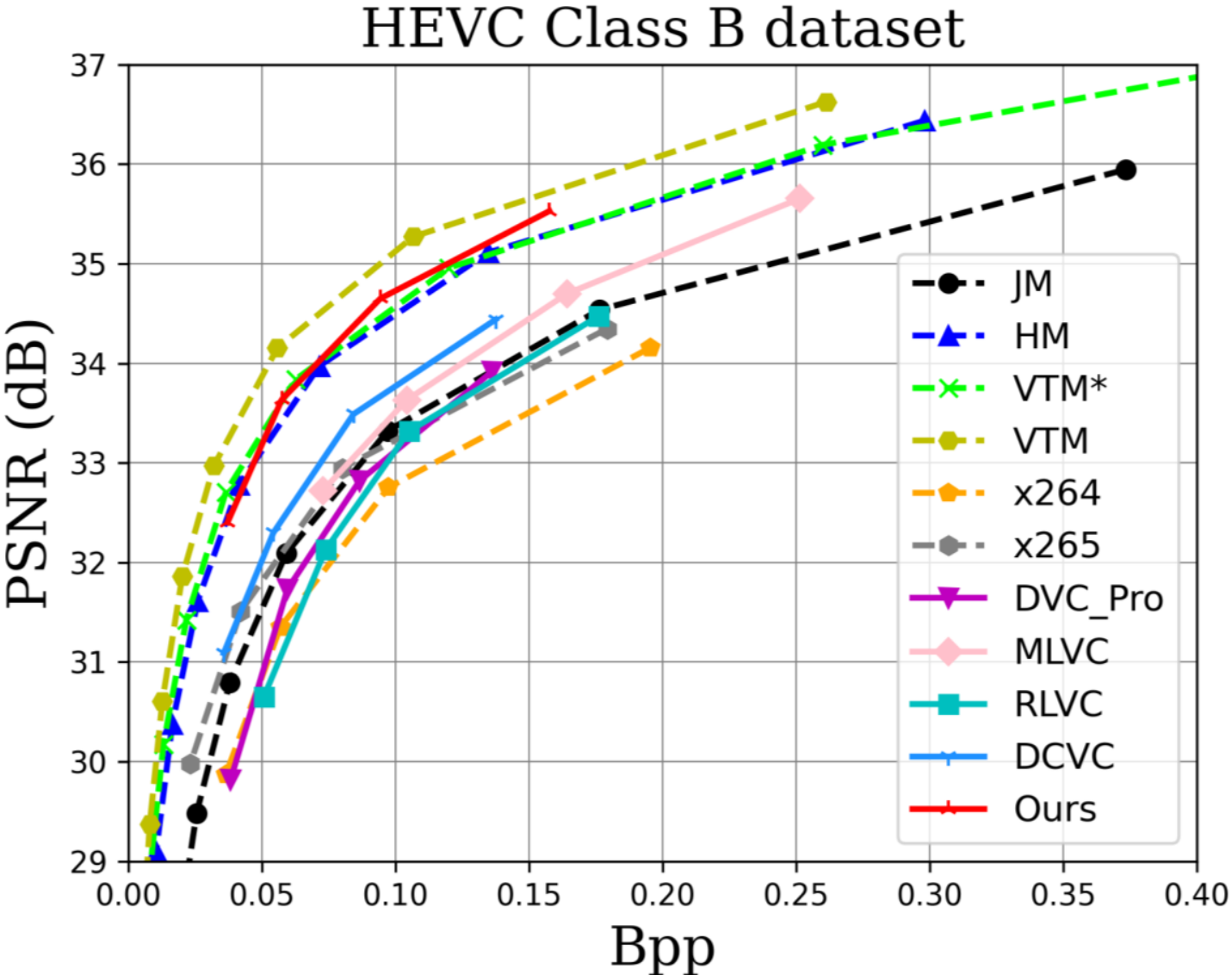}
 \end{minipage}%
  \begin{minipage}[c]{0.29\linewidth}
  \centering
    \includegraphics[width=\linewidth]{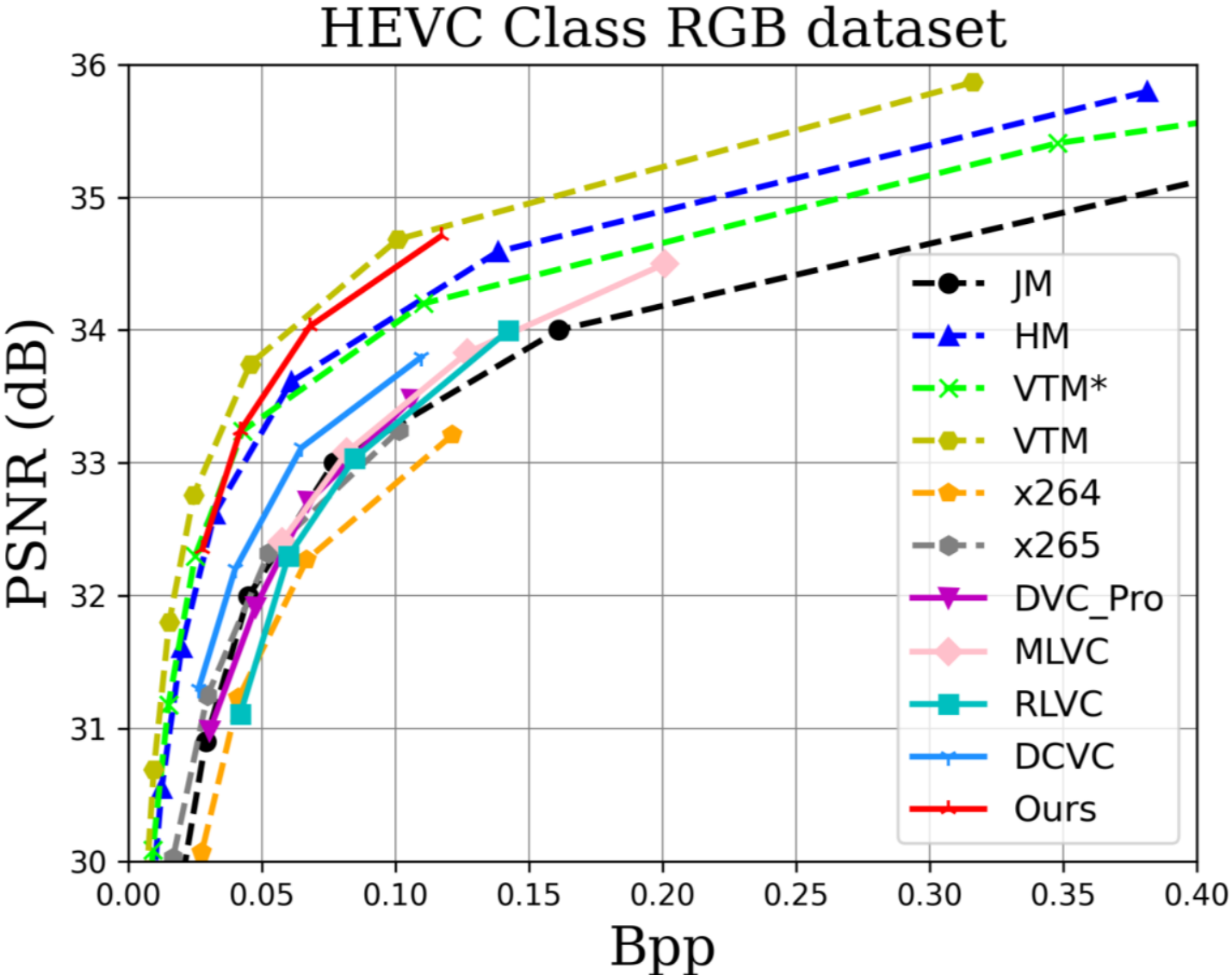}
  \end{minipage}%
  \begin{minipage}[c]{0.29\linewidth}
  \centering
    \includegraphics[width=\linewidth]{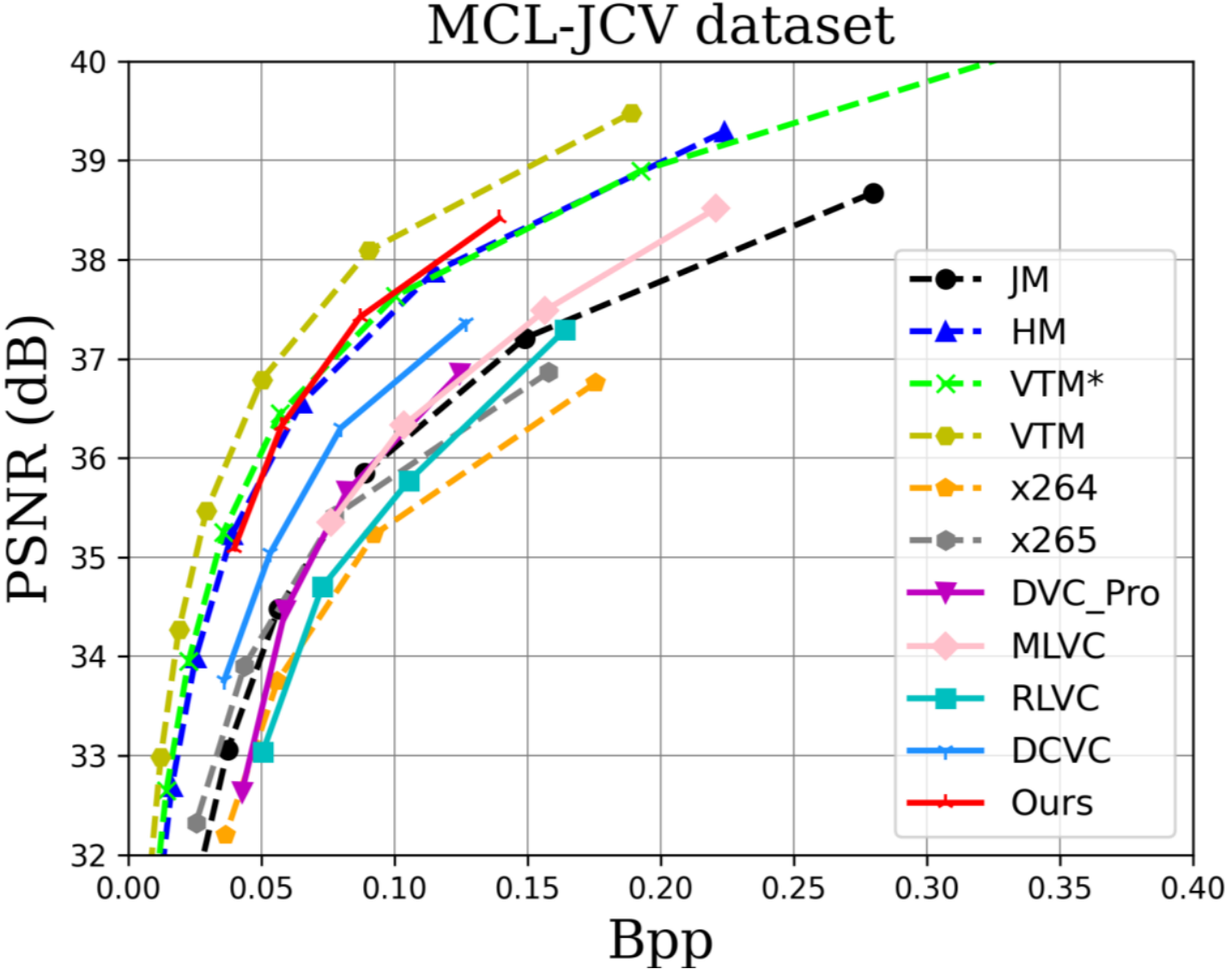}
  \end{minipage}%
  
  \begin{minipage}[c]{0.29\linewidth}
  \centering
    \includegraphics[width=\linewidth]{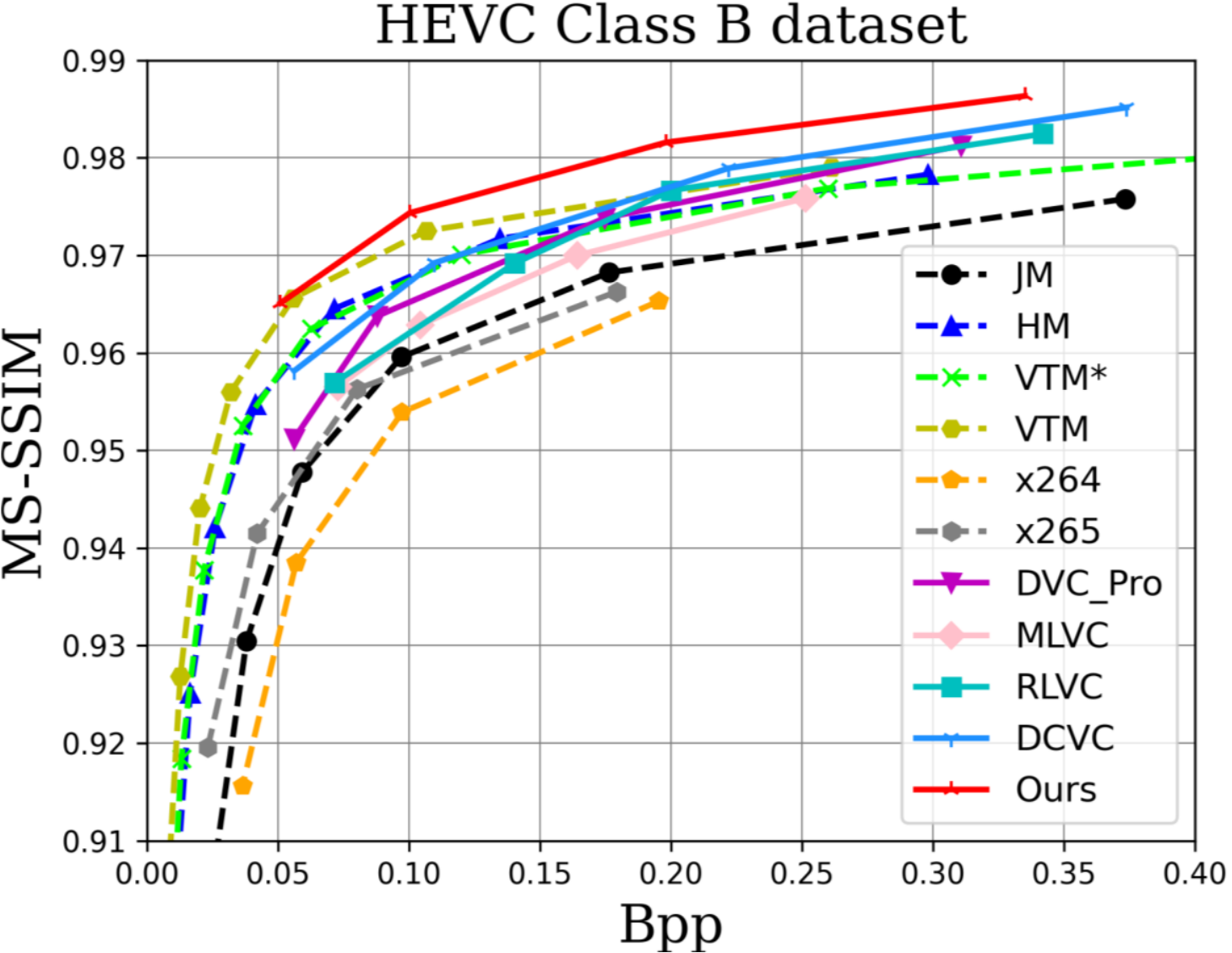}
  \end{minipage}%
  \begin{minipage}[c]{0.29\linewidth}
  \centering
    \includegraphics[width=\linewidth]{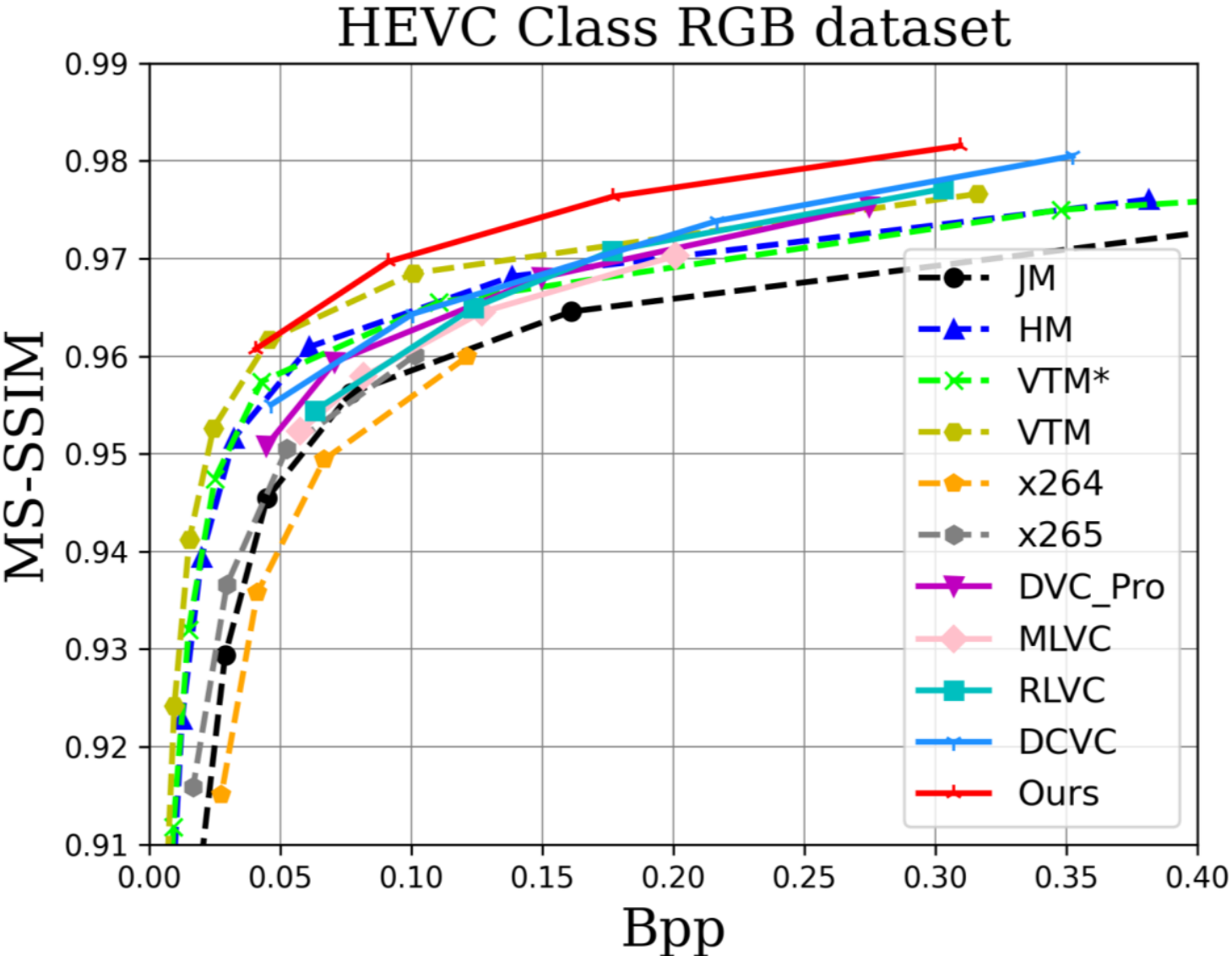}
  \end{minipage}%
  \begin{minipage}[c]{0.29\linewidth}
  \centering
    \includegraphics[width=\linewidth]{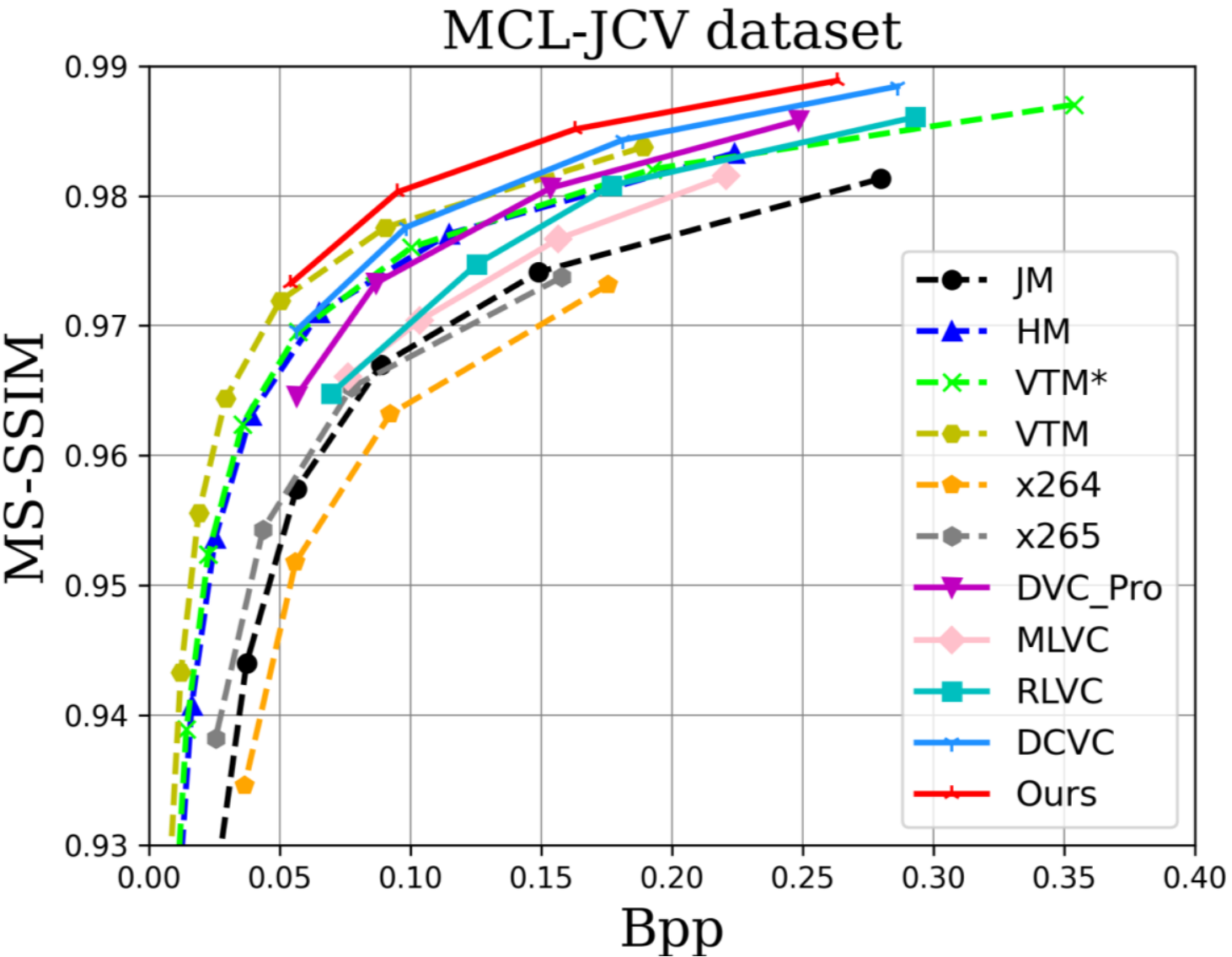}
  \end{minipage}%
    \caption{Rate-distortion performance of our proposed scheme on the HEVC Class B, HEVC Class RGB and MCL-JCV datasets. Intra period is set to 32.}
  \label{fig:result_ip32}
\end{figure*}
\begin{table*}[t]
\caption{BD-rate for PSNR (intra period 32). The anchor is HM.}
  \centering
\scalebox{1}{
\begin{threeparttable}
\begin{tabular}{l|c|c|c|c|c|c|c|c|c|c|c}
\toprule[1.5pt]
               & HM  & JM & VTM &VTM* & x264   & x265  & DVC\_Pro & MLVC  & RLVC  & DCVC  & Ours  \\ \hline
UVG            & 0.0 &108.1&--28.9&4.6& 176.8 & 109.2 & 137.7  & 66.5  & 140.1 & 67.3  & {\bf--9.0} \\ \hline
MCL-JCV     & 0.0 &95.4&--31.2&--7.2&  143.3 & 84.4  & 99.3    & 66.8  & 124.8  & 42.8  &  {\bf--3.2} \\ \hline
MCL-JCV-26     & 0.0 &101.9&--31.0&--6.4&161.0 &93.6  & 92.0    & 56.1  & 115.8  & 37.3  &  {\bf--9.7} \\ \hline
HEVC Class B   & 0.0 &96.9&--28.8&--8.2& 144.6 & 76.1  & 123.7   & 61.4  & 122.6 & 56.0  & {\bf--5.3} \\ \hline
HEVC Class C   & 0.0 &56.6&--29.0&--3.8& 79.8  & 46.2  & 124.0   & 124.1 & 118.9 & 76.9  & {\bf15.1} \\ \hline
HEVC Class D   & 0.0 &50.0&--26.5&--3.5& 72.0  & 43.8  & 93.6    & 96.1  & 81.2  & 52.8  & {\bf--5.4} \\ \hline
HEVC Class E   & 0.0 &80.5&--29.1&--10.0& 153.2 & 60.3  & 283.0   & 138.8 & 246.2 & 156.8 & {\bf18.5} \\ \hline
HEVC Class RGB & 0.0 &102.4&--29.7&--1.7& 151.9 & 82.8  & 102.1   & 82.1  & 114.2 & 51.9  & {\bf--14.4}\\ \bottomrule[1.5pt]
\end{tabular}
  \begin{tablenotes}
   \item \footnotesize \dag Unless otherwise specified, we configure JM, HM, and VTM with the highest-compression-ratio settings for low-delay coding.
    \item \footnotesize \ddag VTM* uses one reference frame instead of the default four frames in VTM. Note that our scheme uses one reference frame for motion estimation.
  \end{tablenotes}
\end{threeparttable}}
\label{table:ip32_psnr}
\end{table*}
\begin{table*}[!t]
\caption{BD-rate for MS-SSIM (intra period 32). The anchor is HM.} 
  \centering
\scalebox{1}{
\begin{tabular}{l|c|c|c|c|c|c|c|c|c|c|c}
\toprule[1.5pt]
               & HM  & JM & VTM  &VTM*& x264   & x265  & DVC\_Pro & MLVC  & RLVC  & DCVC & Ours    \\ \hline
UVG            & 0.0 &105.6&--27.0&2.3&169.9 &87.9   &36.2     &64.7   &49.4   &9.2   & {\bf--25.5}\\ \hline
MCL-JCV        & 0.0 &108.5&--30.4&--5.8&141.0 &71.9   &7.8      &50.3   &34.5   &--16.3 & {\bf--38.3}\\ \hline
MCL-JCV-26     & 0.0 &113.4&--30.3&--6.5&152.0 &74.3   &0.7      &44.2   &23.3   &--18.8 & {\bf--40.5}\\ \hline
HEVC Class B   & 0.0 &112.4&--26.9&--4.2&150.3 &71.1   &23.5     &50.2   &28.3   & 0.9  & {\bf--40.8}\\ \hline
HEVC Class C   & 0.0 &61.3&--27.9&--3.4&89.9  &53.5   &17.0     &53.1   &30.0   &--8.9  & {\bf--42.4}\\ \hline
HEVC Class D   & 0.0 &52.4&--25.9&--3.0&80.0  &49.7   &--7.8     &40.4   &0.2    &--24.2 & {\bf--52.6}\\ \hline
HEVC Class E   & 0.0 &90.9&--27.8&--7.8&184.5 &55.0   &110.1    &106.1  &87.1   &38.0  & {\bf--40.9}\\ \hline
HEVC Class RGB & 0.0 &107.4&--27.2&--3.9&135.9 &64.9   &18.3     &51.8   &21.5   &3.3   & {\bf--43.4}\\ \bottomrule[1.5pt]
\end{tabular}}
\label{table:ip32_msssim}
\end{table*}
\begin{figure*}[t]
  \centering
  \begin{minipage}[c]{0.29\linewidth}
  \centering
    \includegraphics[width=\linewidth]{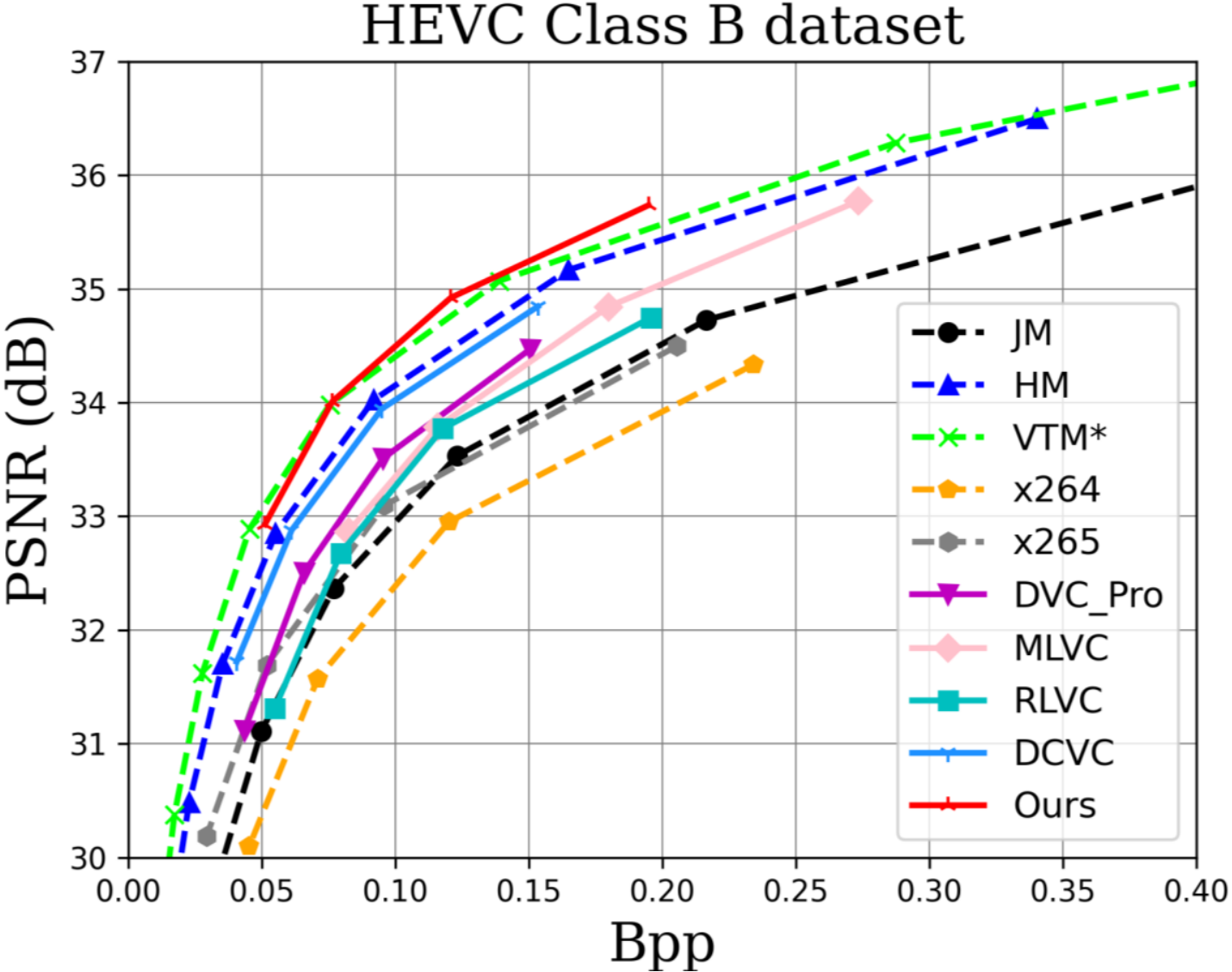}
 \end{minipage}%
  \begin{minipage}[c]{0.29\linewidth}
  \centering
    \includegraphics[width=\linewidth]{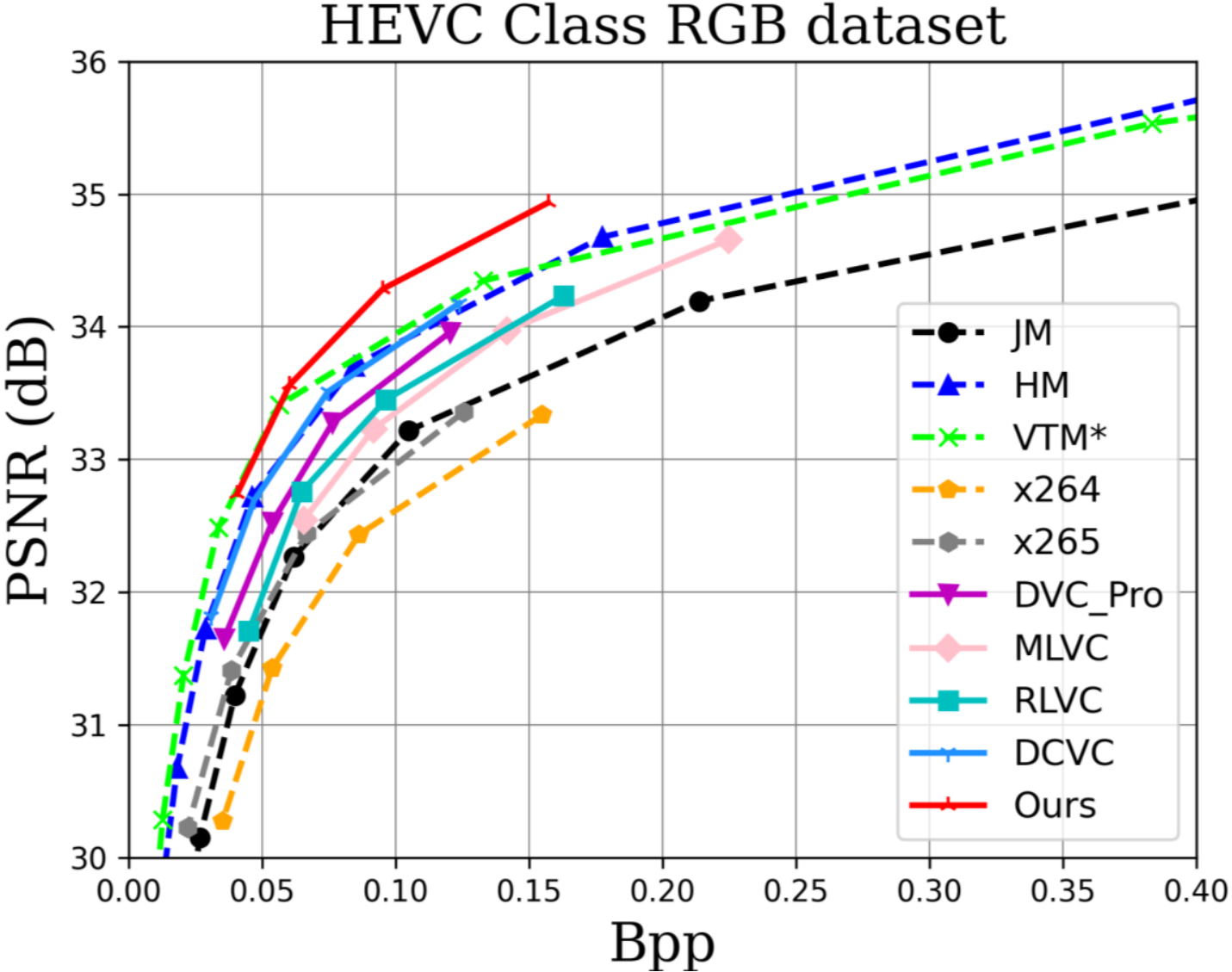}
  \end{minipage}%
  \begin{minipage}[c]{0.29\linewidth}
  \centering
    \includegraphics[width=\linewidth]{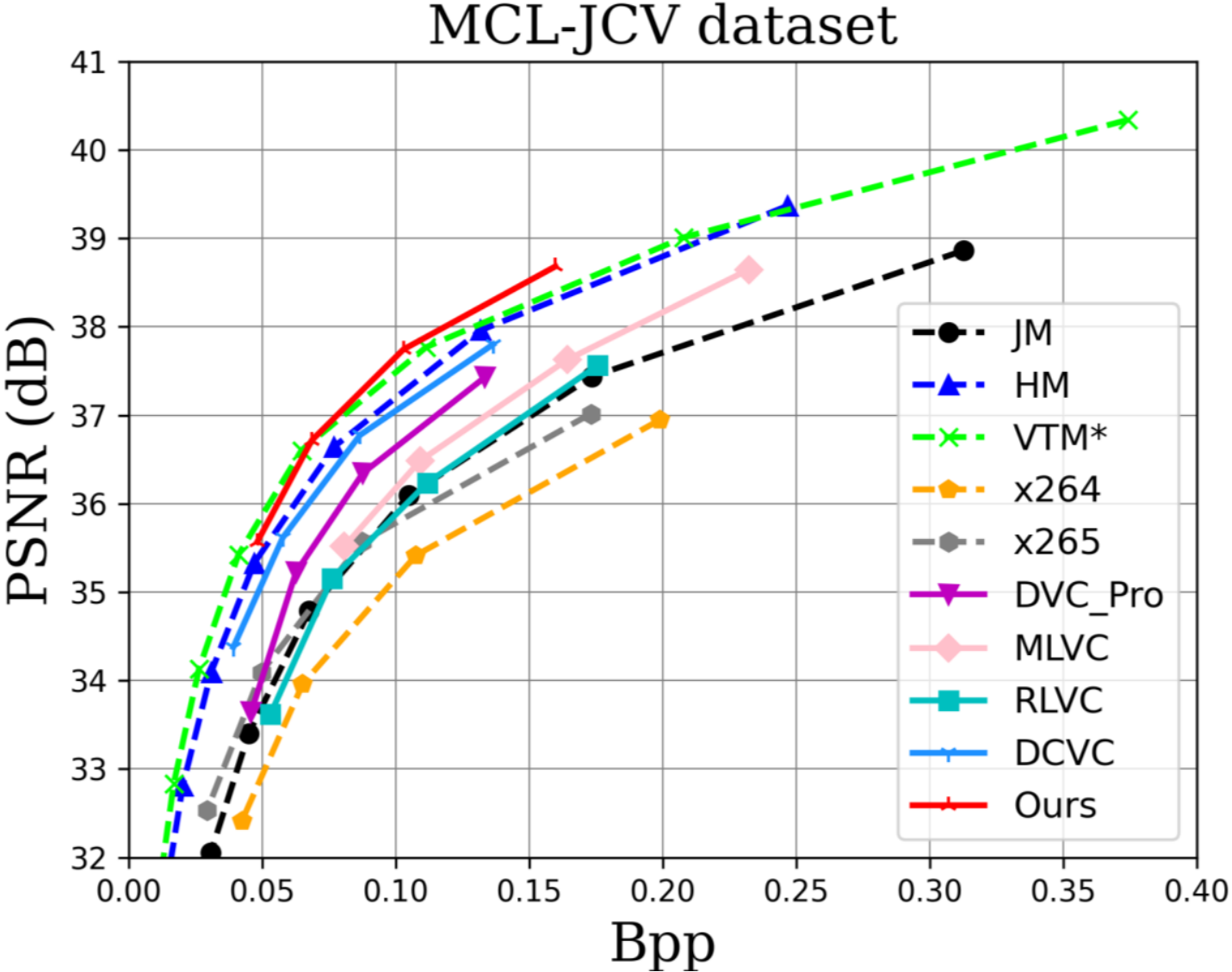}
  \end{minipage}%
  
  \begin{minipage}[c]{0.29\linewidth}
  \centering
    \includegraphics[width=\linewidth]{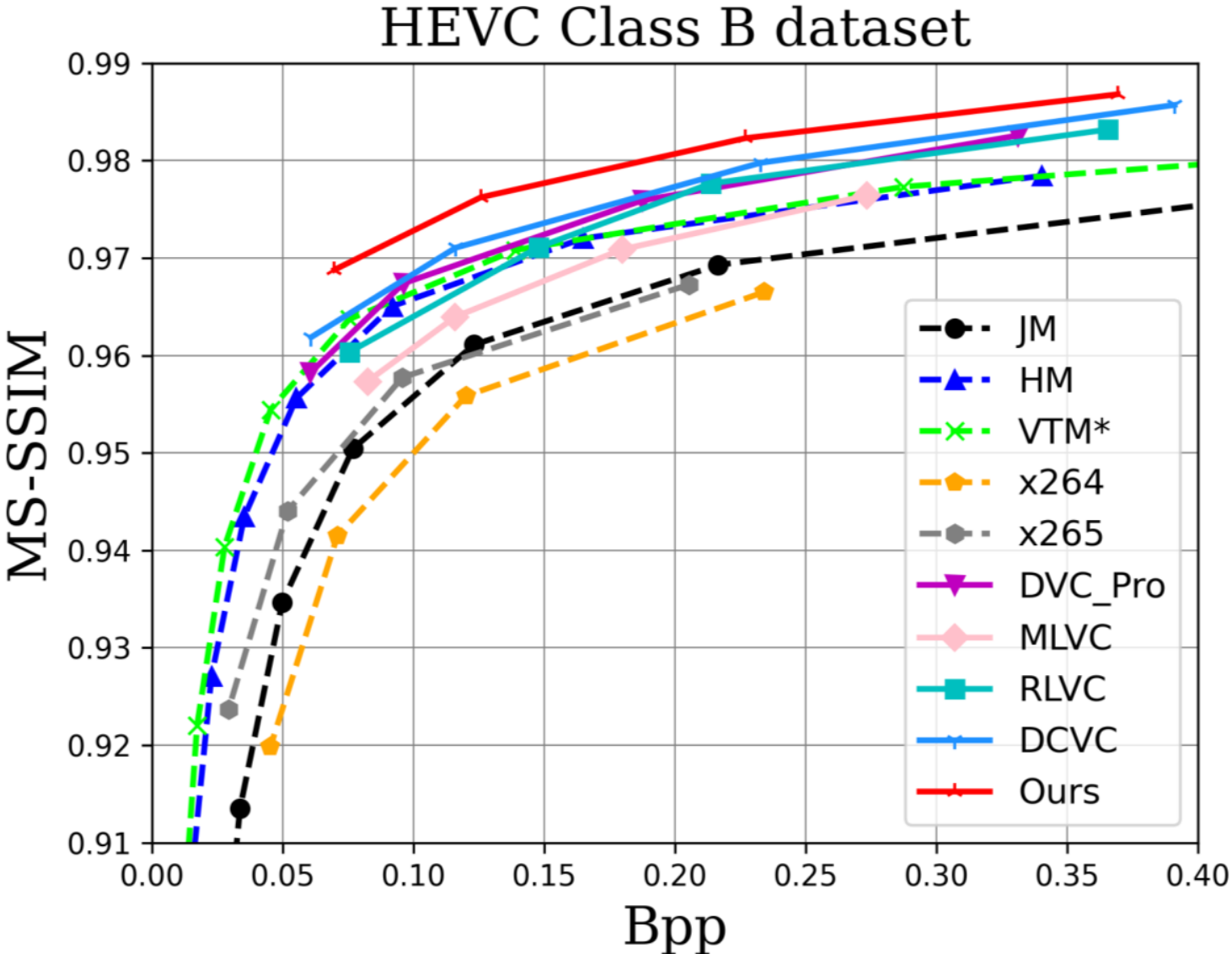}
  \end{minipage}%
  \begin{minipage}[c]{0.29\linewidth}
  \centering
    \includegraphics[width=\linewidth]{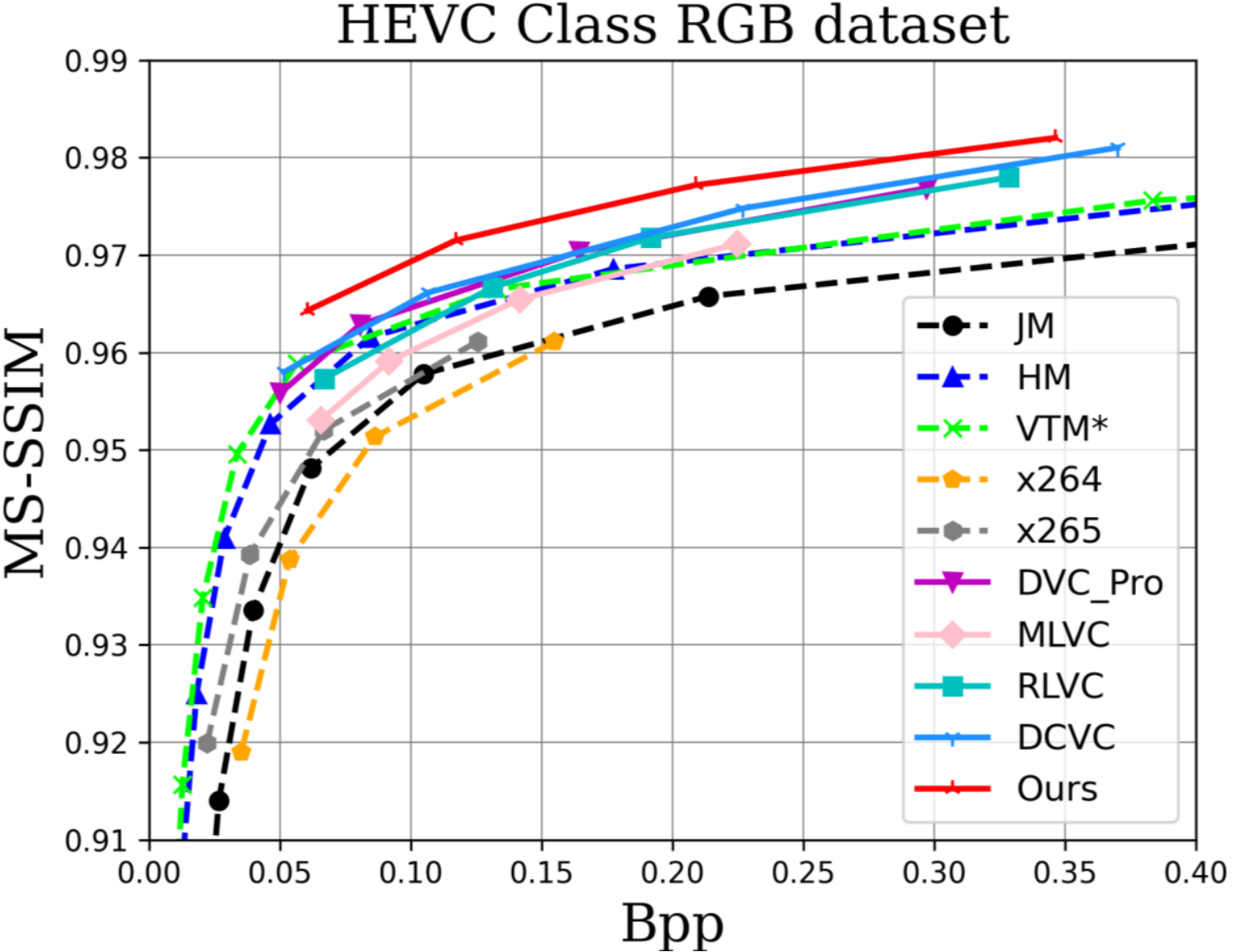}
  \end{minipage}%
  \begin{minipage}[c]{0.29\linewidth}
  \centering
    \includegraphics[width=\linewidth]{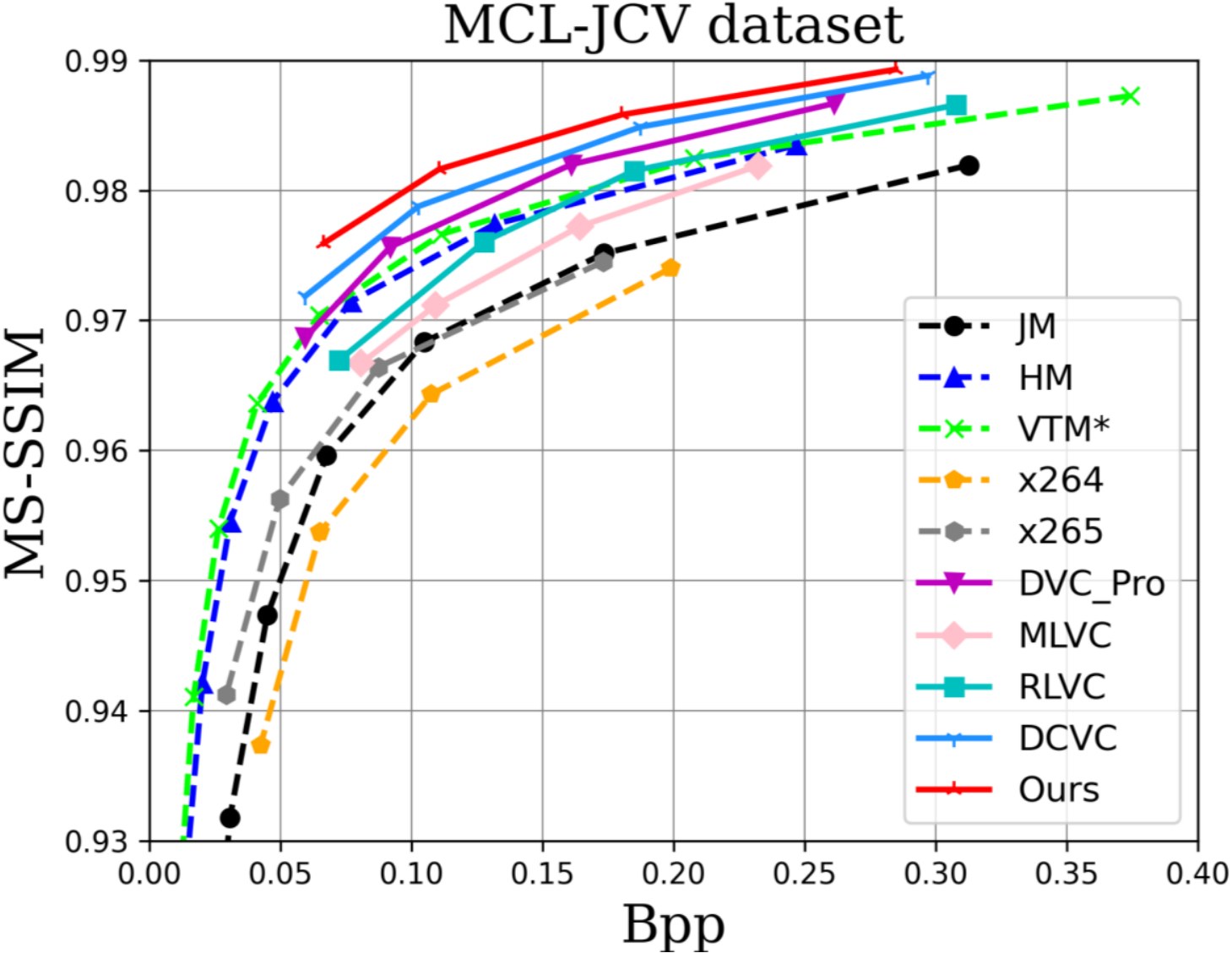}
  \end{minipage}%
    \caption{Rate-distortion performance of our proposed scheme on the HEVC Class B, HEVC Class RGB and MCL-JCV datasets. Intra period is set to 12.}
  \label{fig:result_ip12}
\end{figure*}
\begin{table*}[t]
\caption{BD-rate for PSNR (intra period 12). The anchor is HM.}
  \centering
\scalebox{1}{
\begin{threeparttable}
\begin{tabular}{l|c|c|c|c|c|c|c|c|c|c}
\toprule[1.5pt]
               & HM  & JM &VTM*& x264  & x265 & DVC\_Pro & MLVC & RLVC & DCVC & Ours  \\ \hline
UVG            & 0.0 &79.8&--7.6 &139.2 & 77.7 & 43.5    & 37.8 & 63.2 & 13.8 & {\bf--21.0} \\ \hline
MCL-JCV        & 0.0 &80.0&--13.3&124.4 & 67.8 & 38.7    & 46.4 & 73.6 & 10.0 & {\bf--12.5} \\ \hline
MCL-JCV-26     & 0.0 &84.5&--12.9&138.8 & 74.8 & 32.9    & 37.1 & 65.1 & 4.4 &  {\bf--18.8} \\ \hline
HEVC Class B   & 0.0 &75.8&--16.8 &113.3 & 53.7 & 36.9    & 34.6 & 54.0 & 9.5  & {\bf--15.2} \\ \hline
HEVC Class C   & 0.0 &46.0&--11.7&62.7  & 31.3 & 46.3    & 78.1 & 58.9 & 26.1 & {\bf4.7  } \\ \hline
HEVC Class D   & 0.0 &40.9&--10.4&56.3  & 29.3 & 30.9    & 61.1 & 35.4 & 12.7 & {\bf--10.2} \\ \hline
HEVC Class E   & 0.0 &62.7&--18.3&113.8 & 39.3 & 79.2    & 54.7 & 60.4 & 33.8 & {\bf--6.4 } \\ \hline
HEVC Class RGB & 0.0 &78.7&--11.7&118.8 & 60.2 & 24.2    & 42.4 & 38.7 & 2.2  & {\bf--23.0} \\ \bottomrule[1.5pt]
\end{tabular}
  \begin{tablenotes}
    \item \footnotesize \dag VTM with the highest-compression-ratio settings for low-delay coding is not tested here because it does not support intra period 12.
  \end{tablenotes}
\end{threeparttable}}
\label{table:ip12_psnr}
\end{table*}
\begin{table*}[!htb]
\caption{BD-rate for MS-SSIM (intra period 12). The anchor is HM.}
  \centering
\scalebox{1}{
\begin{tabular}{l|c|c|c|c|c|c|c|c|c|c}
\toprule[1.5pt]
               & HM  & JM & VTM*& x264  & x265 & DVC\_Pro & MLVC & RLVC  & DCVC  & Ours  \\ \hline
UVG            & 0.0 &79.9&--9.3&132.5 & 59.5 & --2.9   & 39.4 & 16.0  & --16.1 & {\bf--35.0} \\ \hline
MCL-JCV     & 0.0 &91.4&--12.3&121.8 & 54.2 & --17.0   & 33.6 & 10.5  & --30.7 & {\bf--44.4} \\ \hline
MCL-JCV-26     & 0.0 &94.7&--13.1&129.8 & 54.9 & --21.3   & 28.5 & 1.6  & --32.8 & {\bf--46.5} \\ \hline
HEVC Class B   & 0.0 &90.2&--13.0&118.6 & 49.4 & --14.7   & 27.0 & --3.7  & --24.2 & {\bf--48.5} \\ \hline
HEVC Class C   & 0.0 &49.7&--11.9&70.6  & 36.7 & --17.5   & 27.3 & --4.4  & --29.8 & {\bf--47.2} \\ \hline
HEVC Class D   & 0.0 &41.8&--9.6&62.2  & 34.9 & --31.4   & 21.2 & --24.7 & --39.7 & {\bf--55.1} \\ \hline
HEVC Class E   & 0.0 &69.7&--14.5&138.1 & 34.6 & 5.0     & 38.8 & --5.3  & --19.5 & {\bf--53.9} \\ \hline
HEVC Class RGB & 0.0 &84.6&--12.8& 103.4 & 43.5 & --20.3  & 20.8 & --11.2 & --25.1 & {\bf--51.4} \\ \bottomrule[1.5pt]
\end{tabular}}
\label{table:ip12_msssim}
\end{table*}
\begin{figure*}[t]
  \centering
   \includegraphics[width=0.85\linewidth]{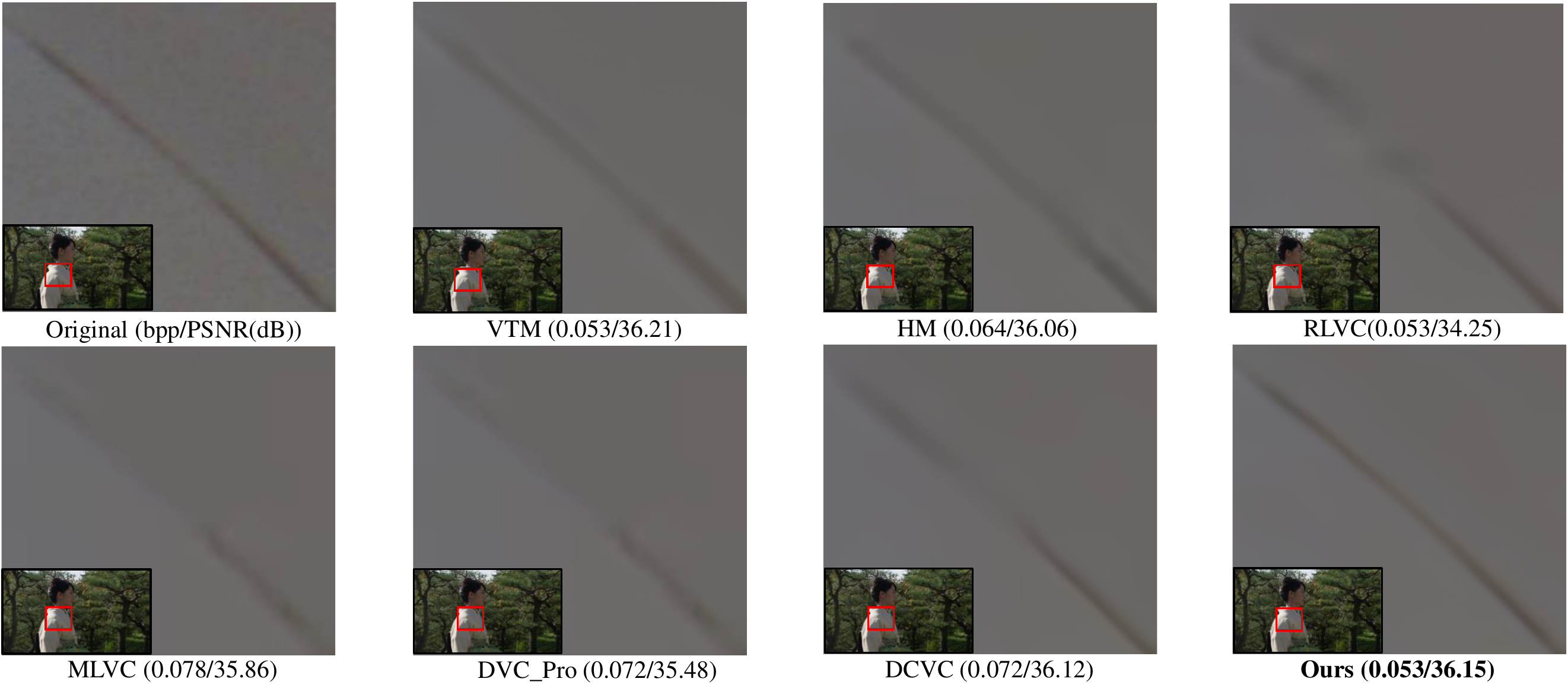}
      \caption{Subjective quality comparison on the 3rd frame of HEVC Class B {\em Kimono} sequence when the intra period is set to 32.}
   \label{fig:subjective}
\end{figure*}
\subsection{Experimental Results}
\label{results}
\subsubsection{ Comparison Setting}
Although random access is the most efficient coding mode, in this paper, we focus on low-delay coding mode following the setting of most previous schemes. Therefore, we set x264, x265, JM-19.0, HM-16.20, and VTM-13.2 under low-delay mode. For x264 and x265, we use the implementation in FFmpeg under {\em veryslow} preset. The internal color space is YUV420 as the previous schemes did.  For JM, HM and VTM, we use the range extension profile instead of the main profile to enable the coding tools designed for non-YUV420.  
Specifically, we use {\em encoder\_JM\_LB\_HE}, {\em encoder\_lowdelay\_main\_rext}, {\em encoder\_lowdelay\_vtm} configuration, respectively. The internal color space is set to YUV444. Meanwhile, four reference frames are used for them as default for seeking the highest compression ratio.
As analyzed in Section~\ref{comparison_with_traditional_video}, the bits of I frames account for a substantial part of the total number of bits. Therefore, for learned video codecs (DVC\_Pro~\cite{lu2020end}, MLVC~\cite{lin2020m}, RLVC~\cite{yang2021learning}, DCVC~\cite{li2021deep}), we use the same image codec as our scheme to compress I frames to make a fair comparison. We adopt the {\em Cheng2020Anchor}~\cite{cheng2020image} implemented by CompressAI~\cite{begaint2020compressai} but we remove its auto-regressive entropy model~\cite{DBLP:conf/nips/MinnenBT18}. The DVC\_Pro models are trained by ourselves and the other models are released from the authors. In previous schemes, the intra period is set to 10 or 12 to limit temporal error propagation. However, such a small intra period is seldom used in real applications, as described in Section~\ref{comparison_with_traditional_video}. Therefore, we propose to use a more reasonable intra period of 32. To provide more information, we also test the cases of intra period 12. Since the GOP size of VTM with {\em encoder\_lowdelay\_vtm} configuration is 8, it does not support intra period 12. Therefore, we add a modified VTM* with low-delay P, 8-bit internal depth, and a single reference frame configuration. We encode 96 frames for each video in all testing datasets.\par
\subsubsection{Results}
Table~\ref{table:ip32_psnr} and Table~\ref{table:ip32_msssim} report the BD-rate~\cite{bjontegaard2001calculation} when the intra period is set to 32. Table~\ref{table:ip12_psnr} and Table~\ref{table:ip12_msssim} report the BD-rate when the intra period is set to 12. The anchor is HM. Negative values indicate bit rate saving compared with HM while positive values indicate bit rate increasing. Experimental results show that, for intra period 32, our scheme outperforms HM by 14.4\% and outperforms VTM* by 14.2\% in terms of PSNR on HEVC Class RGB. Although our scheme still has a 25.4\% PSNR performance gap compared with VTM with the best configuration, we outperform it by 21.1\% in terms of MS-SSIM. For intra period 12, we achieve a larger compression ratio improvement. Note that, when comparing our scheme with VTM or VTM*, VTM or VTM* are regarded as the anchors and the BD-rate values are re-calculated instead of directly subtracting the values in the aforementioned tables. Compared with the results of intra period 32 and 12, our scheme still achieves a high comparison ratio while that of other learned video codecs is greatly reduced when using a larger intra period. Fig.~\ref{fig:result_ip32} and Fig.~\ref{fig:result_ip12} illustrate the RD-curves for different intra periods on HEVC Class B, HEVC Class RGB, and MCL-JCV. We find that learned video codecs behave better when the source videos are in RGB format. The subjective comparison results illustrated in Fig.~\ref{fig:subjective} show that our scheme can retain more details. \par
\begin{table}[t]
 \centering
 \caption{Average encoding/decoding time for a 1080p frame (in seconds).}
\scalebox{1}{
\begin{tabular}{c|c|c}
\toprule[1.5pt]
Schemes  & Enc Time & Dec Time            \\ \hline
VTM      & 743.88 s      & 0.31 s        \\ \hline
HM       & 92.58 s       & 0.21 s        \\ \hline
DCVC     & 12.26 s       & 35.59 s        \\ \hline
RLVC     & 72.00 s& 216.67 s              \\ \hline
Ours     & 0.88 s& 0.47 s                 \\ \bottomrule[1.5pt]
\end{tabular}}
\label{time}
\end{table}
\subsubsection{Model Complexity and Encoding/Decoding Time}
The total number of parameters of our scheme is 10.7M. When processing one 1080p frame, the MACs (multiply–accumulate operation) of our scheme is 2.9T, while that of DCVC is 2.4T. Fig.~\ref{fig:decoding_time} illustrates the average encoding and decoding time for one 1080p frame of different codecs. Table~\ref{time} lists the detailed values. Different from existing schemes which excluded the time for entropy coding~\cite{yang2021learning,li2021deep,liu2020conditional} or data transfer between CPU and GPU~\cite{Rippel_2021_ICCV}, we include the time for model inference, entropy modeling, entropy decoding, and data transfer between CPU and GPU. Our encoding time refers to all the time from reading the original frame to writing bitstream into the hard disk and our decoding time refers to all the time from reading bitstream to generating the final reconstructed frame. As traditional video codecs are optimized for CPU, we run them on an Intel(R) Xeon(R) Platinum 8272CL CPU, which is the same with previous schemes\cite{lu2020end,Rippel_2021_ICCV}. For learned video codecs, we run them on a NVIDIA V100 GPU. We only compare with the learned video codecs whose officially released codes enable bitstream writing. The comparison shows that our scheme takes 0.88s to encode a 1080p frame and takes 0.47s to decode a 1080p frame on average. Although the MACs of our scheme increase by about 21\% compared with DCVC, the decoding time decreases by about 98.7\% since we remove the parallelization-unfriendly auto-regressive entropy model, even though it can help improve the compression performance. Compared with HM and VTM, although our scheme achieves less encoding time and similar encoding time, it does so because it is running on a very sophisticated and expensive GPU while HM and VTM are running on a general-purpose CPU. Therefore, the complexity of our scheme, especially the decoding complexity needs to be reduced before it can be used in practice.

\begin{table}[t]
\caption{Effectiveness of different components of our scheme.}
\centering
\scalebox{1}{
\begin{tabular}{c|c|c|c|c|c|c|c}
\toprule[1.5pt]
FP & TCM & TCR &B & C& D& E&RGB\\ \hline
\Checkmark    &\Checkmark    &\Checkmark        & 0.0& 0.0& 0.0& 0.0 & 0.0        \\ \hline
\Checkmark    &\Checkmark    &\XSolidBrush        & 3.5& 4.2&3.7&1.8&1.8        \\ \hline
\XSolidBrush    &\Checkmark    &\Checkmark        &4.9&6.2&5.6&6.5&3.0          \\ \hline
\XSolidBrush    &\Checkmark    &\XSolidBrush        &7.5 &7.7 &7.9&7.5& 4.7       \\ \hline
\Checkmark    &\XSolidBrush  &\XSolidBrush         & 9.4&11.0&12.1&5.0&3.6       \\ \hline
\XSolidBrush  &\XSolidBrush  &\XSolidBrush          & 11.6&13.7&15.2&9.5&5.8       \\ \bottomrule[1.5pt]
\end{tabular}
}
\label{effectiveness}
\end{table}
\begin{table}[t]
\caption{Influence of the Temporal Contexts on Different Components.}
  \centering
\scalebox{1}{
\begin{tabular}{l|c|c|c|c|c}
\toprule[1.5pt]
     & B    & C    & D     & E    & RGB  \\ \hline
w/o contexts in encoder & 28.8 &36.3 &28.6  &14.7  &9.3 \\ \hline     
w/o contexts in decoder & 3.5 & 3.2 & 3.4  & 2.4  &1.5 \\ \hline  
w/o contexts in frame generator & 8.1 & 9.8 & 13.0  & 8.0  & 6.6 \\ \hline 
w/o contexts in entropy model & 5.7 & 5.3 & 5.5  & 7.8  & 4.7 \\ \bottomrule[1.5pt]
\end{tabular}}
\label{table:temporal context influence}
\end{table}

\begin{table}[t]
\caption{Influence of the number of levels of TCM and re-filled temporal contexts.}
  \centering
  \scalebox{1}{
\begin{tabular}{c|c|c|c|c|c}
\toprule[1.5pt]
     & B & C& D& E&RGB  \\ \hline
3L3C & 0.0& 0.0& 0.0& 0.0 & 0.0 \\ \hline
4L1C & 3.6 & 4.4 & 3.6 &1.5 &1.5 \\ \hline
3L1C & 3.5 & 4.2 & 3.7 &1.8 &1.8 \\ \hline
2L1C & 5.4 & 6.4 & 5.8 &3.6 &2.7 \\ \hline
1L1C & 9.4 & 11.0&12.1 &5.0 &3.6 \\ \toprule[1.5pt]\bottomrule[1.5pt]
4L4C & 0.7 & 1.2 & 1.0 &0.6 & --0.2 \\ \hline
3L2C & 1.8 & 1.9 & 1.1 &1.0 & 0.4 \\ \hline
2L2C & 3.7 & 4.0 & 3.6 &2.5 &2.2 \\ \bottomrule[1.5pt]
\end{tabular}}
\label{table:ablation1}
\end{table}

\subsection{Ablation Study}
\subsubsection{Effectiveness of the Proposed Different Components}
In our work, we focus on better learning and utilizing temporal contexts. We propose to learn the temporal contexts from the propagated feature and re-fill the learned temporal contexts into the modules of our compression scheme. 
To verify the effectiveness of these ideas, we make an ablation study shown in Table~\ref{effectiveness}, where the baseline is our final solution (i.e., feature propagation (FP) + temporal context mining (TCM) + temporal context re-filling (TCR)). From this table, we can find that learning and re-filling temporal contexts from both reconstructed frame or propagated feature improve the compression ratio. It verifies that the proposed TCM and TCR can better learn and utilize temporal contexts. From Table~\ref{effectiveness}, we also find that the improvement of learning temporal contexts from the propagated features is larger than that from the reconstructed frame. It shows that the propagated feature may contain more temporal information compared with the reconstructed frame. \par
\subsubsection{Influence of the Temporal Contexts on Different Components}
In the temporal context re-filling procedure, we feed the temporal contexts into the contextual encoder-decoder, frame generator, and temporal context encoder to improve the compression performance. We make an ablation study to explore the influence of the temporal contexts on different components. Specifically, we remove the temporal contexts in the encoder, decoder, frame generator, and entropy model, respectively. Table~\ref{table:temporal context influence} shows the BD-rate of the four variants. The baseline is our final solution. We can see that the re-filled temporal context plays an important role in the temporal prediction, frame reconstruction, and temporal entropy modeling.

\begin{table}[t]
\caption{Effectiveness of simply adding more network layers.}
  \centering
  \scalebox{1}{
\begin{tabular}{c|c|c|c|c|c}
\toprule[1.5pt]
           & B &C&D&E&RGB  \\ \hline
3L3C & 0.0& 0.0& 0.0& 0.0 & 0.0 \\ \hline
1L1C & 9.4 & 11.0&12.1 &5.0 &3.6 \\ \hline
1L1C (fe) &7.7 &9.0 &9.9 &4.3 &3.0 \\ \hline
1L1C (cr) &8.1 &8.2 &8.9 &4.9 &2.9 \\ \hline
1L1C (fg) &8.3 &9.5 &9.7 &6.9 &3.5 \\ \bottomrule[1.5pt]
\end{tabular}}
\label{table:1L1C_add_complexity}
\end{table}
\begin{table}[t]
\caption{Influence of the dimension of the propagated feature.}
  \centering
\scalebox{1}{
\begin{tabular}{c|c|c|c|c|c}
\toprule[1.5pt]
     & B    & C    & D     & E    & RGB  \\ \hline
64 channels & 0.0 & 0.0 &0.0  & 0.0  & 0.0 \\ \hline      
48 channels & 1.3 & 1.1 &1.0  & 1.0  & 0.1 \\ \hline    
15 channels & 2.6 & 1.9 & 2.3  & 2.7  & 1.7 \\ \hline
9 channels & 2.5 & 2.6 & 2.7  & 2.7  & 1.8 \\ \bottomrule[1.5pt]
\end{tabular}}
\label{table:DPB}
\end{table}
\subsubsection{Influence of the Number of Levels of TCM and Re-filled Temporal Contexts}
To study the influence of the number of levels of TCM module, we change the number of levels of the hierarchical structure of TCM but only output a single scale temporal context $\bar{C}_{t}^{0}$, e.g., only $\bar{C}_{t}^{0}$ is utilized in the TCR procedure. $nL1C$ refers to the number of levels of TCM but only outputs a single scale context. The baseline is our final solution. As shown in Table~\ref{table:ablation1}, the bit rate saving is improved as the level of the TCM module increases but begins to saturate when the number of levels is 3. \par
To study the influence of the number of re-filled temporal contexts, we change the number of the output contexts of TCM. Table~\ref{table:ablation1} shows that the bit rate saving is improved as the number of re-filled temporal contexts increases. When the number of re-filled temporal contexts exceeds 3, the performance improvement is not obvious. In this paper, we employ 3 temporal contexts.\par
To demonstrate that the compression ratio improvement is not simply owing to using more network layers, we add more residual blocks to the feature extraction ($fe$) module in Eq.~\eqref{equ1}, the context refinement ($cr$) module in Eq.~\eqref{equ5}, and the frame generator ($fg$) in Fig.~\ref{fig:framework} of $1L1C$ model to make the MACs comparable with our final solution. We refer to the modified models as $1L1C(fe)$, $1L1C(cr)$, and $1L1C(fg)$, respectively. Table~\ref{table:1L1C_add_complexity} shows that simply adding network layers cannot improve the compression ratio greatly. \par
\subsubsection{Influence of the Dimension of Propagated Features}
For a fair comparison, we choose 64 as the feature dimension to have the same DPB size as RLVC [52].  To explore the influence of the dimension of propagated features, we also change the feature dimension to 48 (same DPB size with FVC [22]), 15 (same DPB size with MLVC [24]), and 9 (same DPB size with 4 reference frame codecs).
Table~\ref{table:DPB} shows that a low dimension only brings a slight loss.  Users can adaptively adjust the dimension according to hardware capability. \par
\begin{figure}
\begin{center}
   \subfigure[Intra Period 32]{
  \includegraphics[width=0.45\linewidth]{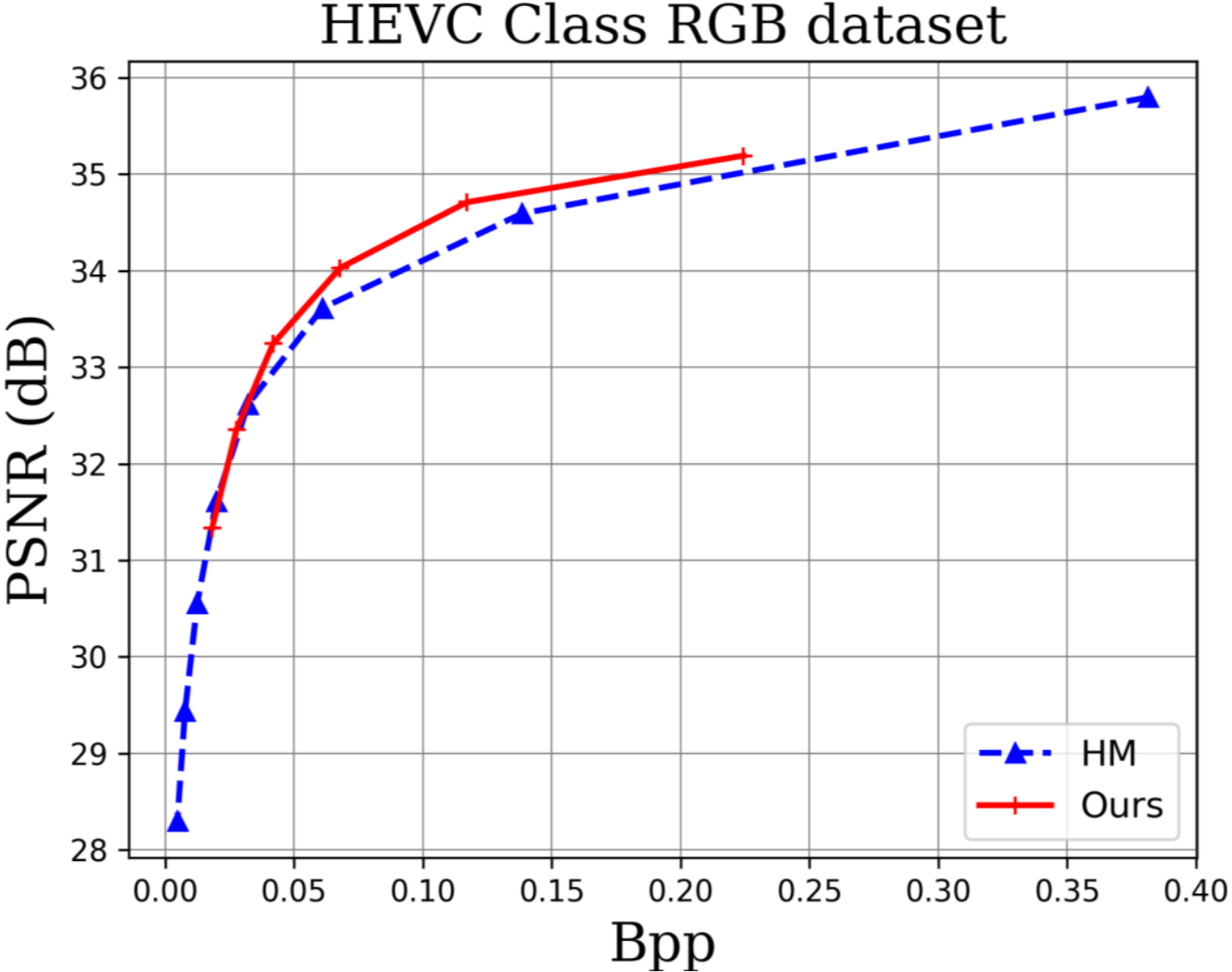}
  }
   \subfigure[Intra Period 12]{
  \includegraphics[width=0.445\linewidth]{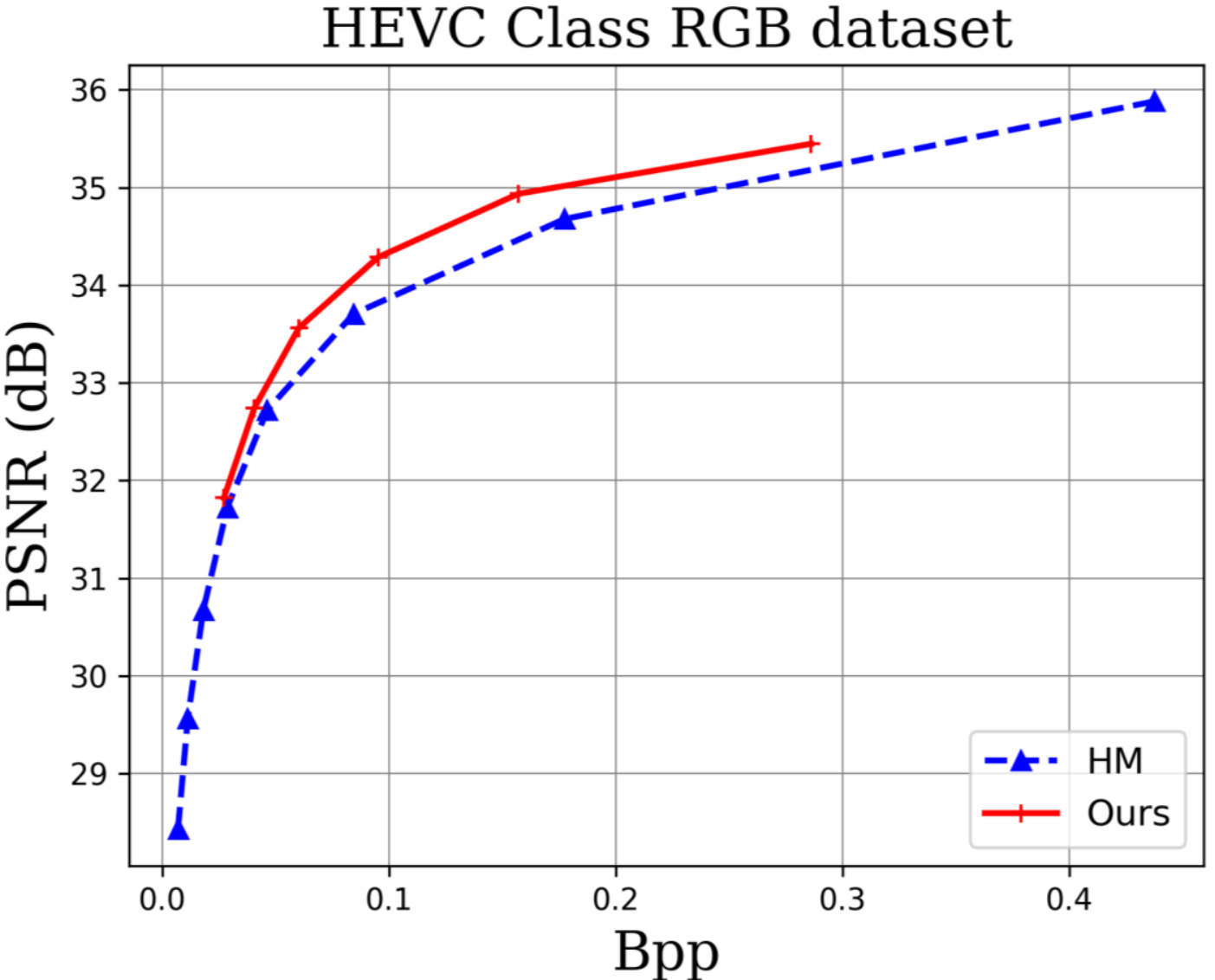}
  }
\end{center}
	\caption{Rate-distortion performance of our proposed scheme on the HEVC Class RGB dataset when the bit rate is extended to lower and higher. }
  \label{bit_rate_range}
\end{figure}
\subsubsection{Wider Bit Rate Range}
It is a common practice to train four models with different bit rate in previous learned video codecs. To make it easier to compare with the previous learned video codecs, our proposed scheme follows the same setting. However, it is still necessary to explore whether the learned video codecs are well-performed in band-limited (lower bit rate) and broadband (higher bit rate) environments. Therefore, when oriented to PSNR, we set $\lambda$ to 128 and 4096 to train additional two models. When oriented to MS-SSIM, we set $\lambda$ to 4 and 128. For other learned video codecs (DVC\_Pro~\cite{lu2020end}, MLVC~\cite{lin2020m}, RLVC~\cite{yang2021learning}, DCVC~\cite{li2021deep}), since  their officially released models only support four bit rate points, we can not extend their bit rate ranges. Fig.~\ref{bit_rate_range} illustrates the rate-distortion performance of our proposed scheme on the HEVC Class RGB  when the bit rate is extended to lower and higher. The results show that our scheme can still work well over a  wider range of bit rates.
\section{Conclusion and Discussion}
\label{sec:conclusion}
In this paper, we propose a temporal context mining module and a temporal context re-filling procedure to better learn and utilize temporal contexts for learned video compression.
Without the auto-regressive entropy model, our proposed scheme achieves higher compression ratio than the existing learned video codecs. Our scheme also outperforms the reference software of H.265/HEVC---HM in terms of PSNR by 14.4\% and outperforms the reference software of H.266/VVC---VTM in terms of MS-SSIM by 21.1\%. Even the proposed scheme outperforms VTM with the highest-compression-ratio settings for low-delay coding in terms of MS-SSIM, it is far behind than VTM in terms of PSNR. In addition, the complexity, especially the decoding complexity still needs to be significantly reduced before it can be used in practice. We will further investigate it in our future work.

\bibliographystyle{ieeetr}
\bibliography{main}
\begin{IEEEbiography}[{\includegraphics[width=1in,height=1.25in,clip,keepaspectratio]{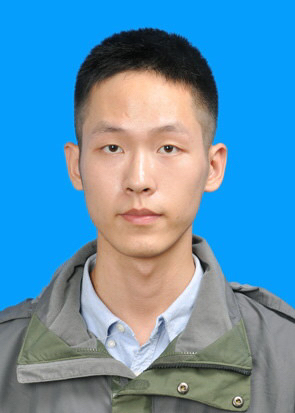}}]{Xihua Sheng}
received the B.S. degree in automation from Northeastern University, Shenyang, China, in 2019. He is currently pursuing the Ph.D. degree in the Department of Electronic Engineering and Information Science at the University of Science and Technology of China, Hefei, China. 
His research interests include image/video/point cloud coding, signal processing, and machine learning.
\end{IEEEbiography}

\begin{IEEEbiography}[{\includegraphics[width=1in,height=1.25in,clip,keepaspectratio]{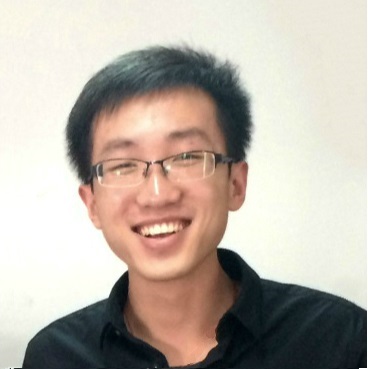}}] {Jiahao Li} received the B.S. degree in computer science and technology from the Harbin Institute of Technology in 2014, and the Ph.D. degree from Peking University in 2019. He is currently a Senior Researcher with the Media Computing Group, Microsoft Research Asia. Previously, he worked on video coding and has more than ten published papers, standard proposals, and patents in this area. His current research interests mainly focus on neural video compression and real-time communication.
\end{IEEEbiography}

\begin{IEEEbiography}[{\includegraphics[width=1in,height=1.25in,clip,keepaspectratio]{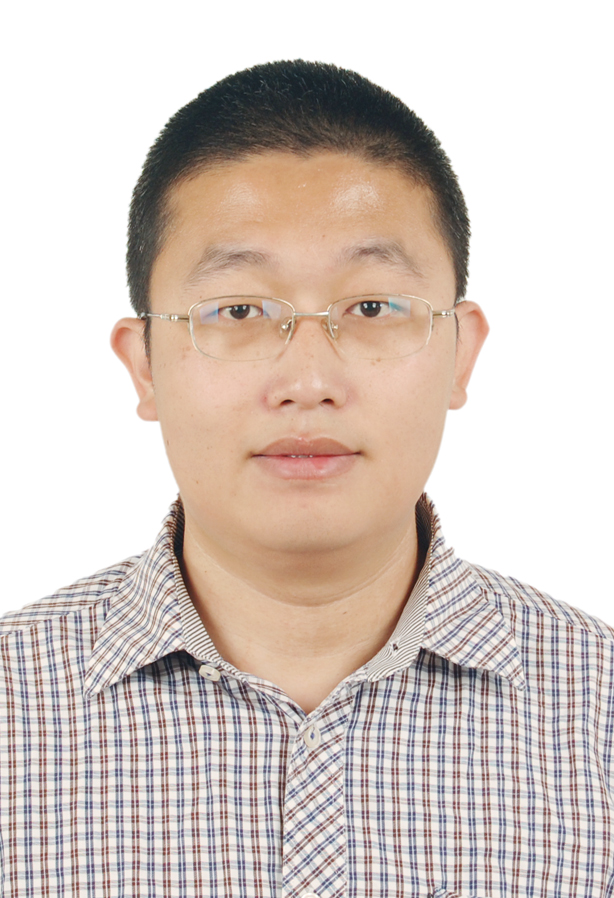}}] {Bin Li} received the B.S. and Ph.D. degrees in electronic engineering from the University of Science and Technology of China (USTC), Hefei, Anhui, China, in 2008 and 2013, respectively.
He joined Microsoft Research Asia (MSRA), Beijing, China, in 2013 and now he is a Principal Researcher. He has authored or co-authored over 50 papers. He holds over 30 granted or pending U.S. patents in the area of image and video coding. He has more than 40 technical proposals that have been adopted by Joint Collaborative Team on Video Coding. His current research interests include video coding, processing, transmission, and communication.\par
Dr. Li received the best paper award for the International Conference on Mobile and Ubiquitous Multimedia from Association for Computing Machinery in 2011. He received the Top 10\% Paper Award of 2014 IEEE International Conference on Image Processing. He received the best paper award of IEEE Visual Communications and Image Processing 2017.
\end{IEEEbiography}

\begin{IEEEbiography}[{\includegraphics[width=1in,height=1.25in,clip,keepaspectratio]{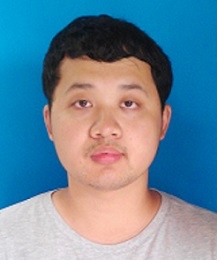}}] {Li Li} (M'17) received the B.S. and Ph.D. degrees in electronic engineering from University of Science and Technology of China (USTC), Hefei, Anhui, China, in 2011 and 2016, respectively.
He was a visiting assistant professor in University of Missouri-Kansas City from 2016 to 2020.
He joined the department of electronic engineering and information science of USTC as a research fellow in 2020 and became a professor in 2022.

His research interests include image/video/point cloud coding and processing.
He received the Best 10\% Paper Award at the 2016 IEEE Visual Communications and Image Processing (VCIP) and the 2019 IEEE International Conference on Image Processing (ICIP).
\end{IEEEbiography}

\begin{IEEEbiography}[{\includegraphics[width=1in,height=1.25in,clip,keepaspectratio]{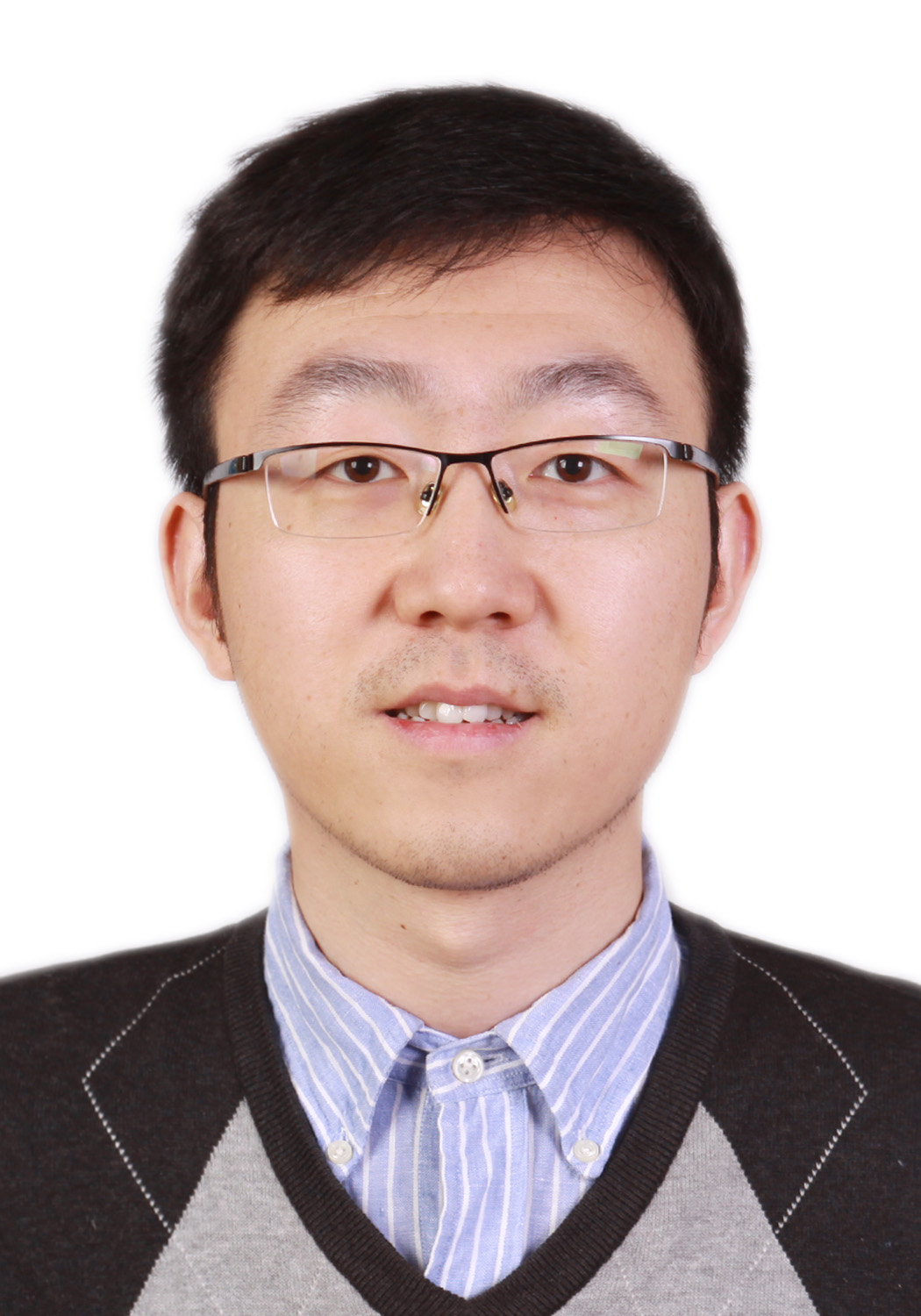}}]{Dong Liu}
(M'13--SM'19) received the B.S. and Ph.D. degrees in electrical engineering from the University of Science and Technology of China (USTC), Hefei, China, in 2004 and 2009, respectively. He was a Member of Research Staff with Nokia Research Center, Beijing, China, from 2009 to 2012. He joined USTC as a faculty member in 2012 and became a Professor in 2020.

His research interests include image and video processing, coding, analysis, and data mining.
He has authored or co-authored more than 200 papers in international journals and conferences. He has more than 20 granted patents. He has several technical proposals adopted by international or domestic standardization groups.
He received the 2009 \textsc{IEEE Transactions on Circuits and Systems for Video Technology} Best Paper Award and the VCIP 2016 Best 10\% Paper Award. He and his students were winners of several technical challenges held in ISCAS 2022, ICCV 2019, ACM MM 2019, ACM MM 2018, ECCV 2018, CVPR 2018, and ICME 2016. He is a Senior Member of CCF and CSIG, an elected member of MSA-TC of IEEE CAS Society. He serves or had served as the Chair of IEEE 1857.11 Standard Working Subgroup, a Guest Editor for \textsc{IEEE Transactions on Circuits and Systems for Video Technology}, an Associate Editor for \emph{Frontiers in Signal Processing}, an Organizing Committee member for VCIP 2022, ICME 2021, ICME 2019, etc.
\end{IEEEbiography}

\begin{IEEEbiography}[{\includegraphics[width=1in,height=1.25in,clip,keepaspectratio]{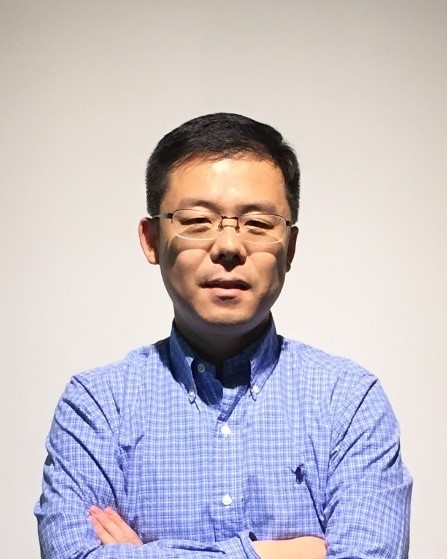}}] {Yan Lu} received his Ph.D. degree in computer science from Harbin Institute of Technology, China. He joined Microsoft Research Asia in 2004, where he is now a Partner Research Manager and manages research on media computing and communication. He and his team have transferred many key technologies and research prototypes to Microsoft products. From 2001 to 2004, he was a team lead of video coding group in the JDL Lab, Institute of Computing Technology, China. From 1999 to 2000, he was with the City University of Hong Kong as a research assistant. Yan Lu has broad research interests in the fields of real-time communication, computer vision, video analytics, audio enhancement, virtualization, and mobile-cloud computing. He holds 30+ granted US patents and has published 100+ papers in refereed journals and conference proceedings.
\end{IEEEbiography}

\end{document}